%% file: main.tex
\documentclass[5p, times]{elsarticle}

\usepackage[T1]{fontenc}
\usepackage[utf8]{inputenc}
\usepackage{booktabs}

\usepackage{graphicx} 
\usepackage{bbding} 
\usepackage{amsmath}
\usepackage{amssymb} 
\usepackage{algorithm,algpseudocode}

\usepackage{xparse}
\usepackage{morewrites}
\usepackage{wrapfig}
\usepackage{xkeyval}
\usepackage{balance}

\usepackage{siunitx} 
\providecommand{\etal}{\textit{et~al.}}

\usepackage[dvipsnames]{xcolor}
\usepackage[colorlinks,bookmarksopen,bookmarksnumbered,citecolor=Blue,linkcolor=Blue,urlcolor=red]{hyperref}
\usepackage[capitalize,nameinlink,noabbrev]{cleveref} % Needs to be included after hyperref!

\newwrite\authorbibfile%

\AtBeginDocument{%
  \immediate\openout\authorbibfile=\jobname.aub%
}%

\AtEndDocument{%
\immediate\closeout\authorbibfile
\InputIfFileExists{\jobname.aub}{}{}
}%

\makeatletter
\define@key{authorbib}{scale}[1]{%
\def\AuthorbibKVMacroScale{#1}%
}

\define@key{authorbib}{wraplines}[10]{%
\def\AuthorbibKVMacroWraplines{#1}%
}

\define@key{authorbib}{imagewidth}[4cm]{%
\def\AuthorbibKVMacroImagewidth{#1}%
}

\define@key{authorbib}{overhang}[10pt]{%
\def\AuthorbibKVMacroOverhang{#1}%
}

\define@key{authorbib}{imagepos}[l]{%
\def\AuthorbibKVMacroImagepos{#1}%
}

 \def\ps@pprintTitle{%
     \let\@oddhead\@empty
     \let\@evenhead\@empty
     \def\@oddfoot{\centerline{\textit{\small{Preprint submitted to a journal}}}}
     \let\@evenfoot\@oddfoot}
 
\makeatother

\presetkeys{authorbib}{imagepos=l,imagewidth=4cm,wraplines=15,overhang=20pt}{}

\newlength{\AuthorbibTopSkip}
\newlength{\AuthorbibBottomSkip}
\NewDocumentCommand{\authorbibliography}{+o+m+m+m}{%
  \IfNoValueTF{#1}{%
  }{%
    \setkeys{authorbib}{#1}%
    \immediate\write\authorbibfile{%
      \string\begin{wrapfigure}[\AuthorbibKVMacroWraplines]{\AuthorbibKVMacroImagepos}[\AuthorbibKVMacroOverhang]{\AuthorbibKVMacroImagewidth}^^J
        \string\includegraphics[width=2.5cm, keepaspectratio]{#2}^^J
        \string\end{wrapfigure}^^J
    }%
  }%
  \IfNoValueTF{#3}{%
    \typeout{Warning: No author name}%
  }{%
    \immediate\write\authorbibfile{%
      \unexpanded{\vspace{\AuthorbibTopSkip}}^^J
      \string\noindent\relax
      \unexpanded{\textbf{#3}} %\par}^^J
      \unexpanded{#4}^^J%
      \unexpanded{\vspace{\AuthorbibBottomSkip}}^^J
      }%
  }%
}%

\begin{document}

\begin{frontmatter}

\title{Stereo obstacle detection for unmanned surface vehicles by IMU-assisted semantic segmentation}

\author[fri]{Borja~Bovcon\corref{cor1}}
\ead{borja.bovcon@fri.uni-lj.si}

\author[fri,fe]{Rok~Mandeljc}
\ead{rok.mandeljc@fe.uni-lj.si}

\author[fe]{Janez~Perš}
\ead{janez.pers@fe.uni-lj.si}

\author[fri]{Matej~Kristan}
\ead{matej.kristan@fri.uni-lj.si}

\address[fri]{University of Ljubljana, Faculty of Computer and Information Science, Večna pot 113, 1000 Ljubljana, Slovenia.}
\address[fe]{University of Ljubljana, Faculty of Electrical Engineering, Tržaška cesta 25, 1000 Ljubljana, Slovenia.}

\cortext[cor1]{Corresponding author}

%%%%%%%%%%%%%%%%%%%%%%%%%%%%%%%
%%         Abstract          %%
%%%%%%%%%%%%%%%%%%%%%%%%%%%%%%%
\begin{abstract}
A new obstacle detection algorithm for unmanned surface vehicles (USVs) is presented. A state-of-the-art graphical model for semantic segmentation is extended to incorporate boat pitch and roll measurements from the on-board inertial measurement unit (IMU), and a stereo verification algorithm that consolidates tentative detections obtained from the segmentation is proposed. 
The IMU readings are used to estimate the location of horizon line in the image, which automatically adjusts the priors in the probabilistic semantic segmentation model. 
We derive the equations for projecting the horizon into images, propose an efficient optimization algorithm for the extended graphical model, and offer a practical IMU-camera-USV
calibration procedure. Using an USV equipped with multiple synchronized sensors, we captured a new challenging multi-modal dataset, and annotated its images with water edge
and obstacles. Experimental results show that the proposed algorithm significantly outperforms the state of the art, with nearly \SI{30}{\percent} improvement in water-edge detection accuracy, 
an over \SI{21}{\percent} reduction of false positive rate, an almost \SI{60}{\percent} reduction of false negative rate, and an over \SI{65}{\percent} increase of true positive rate, while its Matlab
implementation runs in real-time.
\end{abstract}

\end{frontmatter}

%%%%%%%%%%%%%%%%%%%%%%%%%%%%%%%
%%       Introduction        %%
%%%%%%%%%%%%%%%%%%%%%%%%%%%%%%%
\section{Introduction}
\label{sec:intro}
The past decade of research in marine and field robotics has led to establishment of a new class of small-sized unmanned surface vehicles (USVs). Such boats are typically less than \SI{2}{\meter} long, 
and can be guided either manually or programmed to follow a pre-determined path. Their main advantage over the larger counterparts is portability and the ability to navigate relatively shallow waters and 
narrow marinas. This broadens the potential areas of applications, which range from coastal water and environmental surveillance to inspection of man-made structures above and below water.

On the other hand, the small form factor of such USVs limits the available sensor payload, which is further restricted by the power consumption limitations. Therefore, cameras are gaining prominence as 
light-weight, low-power, and information-rich sensors, which represent a viable alternative or addition to other sensor modalities~\cite{Woo2016,Prasad2017}. In contrast to LIDAR, the camera systems do 
not contain moving parts, which makes them robust to mechanical stress. Compared to RADAR and LIDAR, the camera systems are completely passive and are therefore inherently safe in any environment, without 
the potential risk of interfering with other critical systems, such as radio communication or GPS~\cite{Halterman2010}. Furthermore, the commercially available high-performance LIDAR systems are typically
unsuitable for small-sized USVs, as their weight and size may compromise the boat stability~\cite{Muhammad2010}. Similarly, the usefulness of affordable and commonly used laser sensors is limited
even on larger vessels, due to significant variations in lighting and constant rocking of the boat~\cite{Dahlkamp2006,Rankin2010}. 

\begin{figure}[h!]
	\centering
	\includegraphics[width=1\columnwidth]{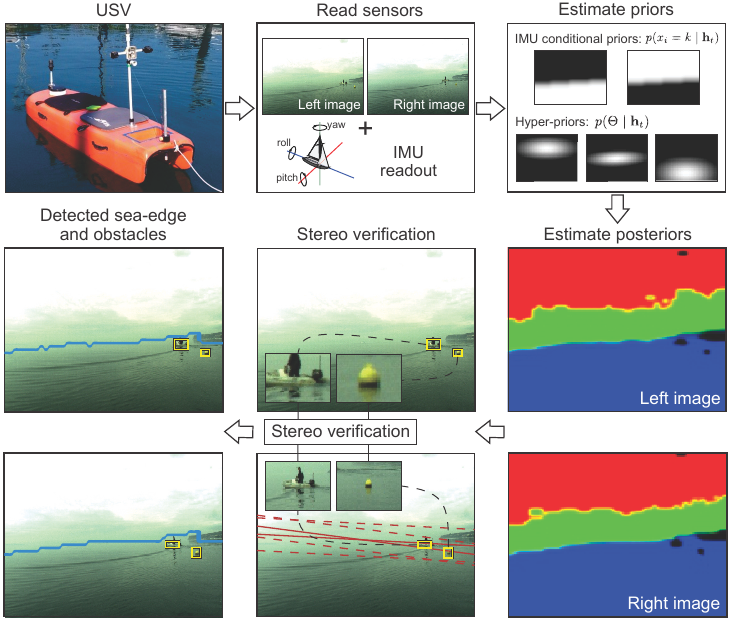}
	\caption{Outline of our IMU-assisted semantic segmentation method using stereo verification (ISSM\textsubscript{S}) for obstacle detection. The IMU measurements constrain the semantic segmentation, and provide sea-edge estimate and obstacle candidates in each camera. Afterwards, the stereo verification step is applied to reduce the false negative and false positive detections.}%
	\label{fig:our_approach_stereo}
\end{figure}

The performance of standard stereo-vision-based methods in USV applications is severely limited by rapidly-changing water surface, reflections, and the absence of texture in the absence of disturbances, such as waves. The obstacles often 
do not sufficiently protrude through the water surface to be reliably detected by stereo systems. Recently, Kristan~\etal~\cite{KristanCYB2015} proposed a graphical model for monocular obstacle detection via semantic 
segmentation (SSM) of the observed marine scene. The algorithm generates a water segmentation mask and treats all objects in the water as obstacles. Their approach runs in real-time and outperforms related approaches
on the task of obstacle detection. However, it fails in the presence of significant rolling and pitching in rough seas, and is susceptible to false positives and degraded sea-edge estimation in the presence of visual ambiguities (e.g., 
when horizon is poorly distinguishable from the sky).

In this paper, we build upon~\cite{KristanCYB2015}, and propose a new seg\-men\-ta\-ti\-on-based obstacle detector for unmanned surface vehicles that incorporates the roll and pitch measurements from the on-board IMU and additionally 
verifies obtained detections using a stereo system (\cref{fig:our_approach_stereo}). We claim the following three major contributions. The first contribution is extension of the graphical model for semantic 
segmentation~\cite{KristanCYB2015} with roll and pitch measurements from the IMU. This information is used to project the horizon onto the input image, and automatically adjust the priors and hyper-priors of the segmentation model. This enables reliable segmentation even during significant motion of the boat. We derive the required IMU-to-camera horizon projection equations, and propose a practical IMU-to-camera calibration procedure. Our second contribution is improvement of the seg\-men\-ta\-ti\-on-based obstacle detection via 
stereo verification. The approach applies epipolar constraints and template matching to reduce the amount of false positive and false negative detections. Our third contribution is a new challenging dataset for marine 
obstacle detection, which consists of multiple sequences with time-synchronized data streams from on-board stereo system, IMU and GPS, and is currently the largest of its kind. The extensive experimental analysis on 
this new dataset shows that the proposed approach significantly outperforms the state-of-the-art SSM~\cite{KristanCYB2015}, both in accuracy of sea-edge estimation and in the accuracy of obstacle detection.

The remainder of this paper is structured as follows. \Cref{sec:relatedWork} provides a brief overview of the related work. \Cref{sec:algDesc,sec:stereoDetection,sec:cameraImuCalibration} present the proposed IMU-assisted 
segmentation model, stereo-based obstacle verification algorithm, and calibration procedure, respectively. \Cref{sec:dataset} describes our new marine obstacle detection dataset, which is used in \cref{sec:experiments} 
for experimental evaluation of the proposed approach, while \cref{sec:conclusionAndFutureWork} wraps up the paper with concluding remarks.

%%%%%%%%%%%%%%%%%%%%%%%%%%%%%%%
%%  Related work description %%
%%%%%%%%%%%%%%%%%%%%%%%%%%%%%%%
\section{Related work} \label{sec:relatedWork}
Obstacle detection in unmanned surface vehicles is still a relatively young research area, especially compared to the already-established field of autonomous ground vehicles, where a considerable body of literature can be found on topic of obstacle detection and avoidance. Krotosky~\etal~\cite{krotosky2007color}, Zhang~\etal~\cite{zhang2015novel}, and Cao~\etal~\cite{cao2015perception} use a stereo camera system to compute disparity map and use it for obstacle detection.
They apply different computer vision methods to filter the disparity map and remove noise in detected obstacles. 
Krotosky~\etal~\cite{krotosky2007color} combine detections from disparity map with information from an infrared camera to further improve the detection of pedestrians. 
Shim~\etal~\cite{shim2015autonomous} use a combination of several sensors for obstacle detection. They use LIDAR to detect general obstacles on the road under the assumption of ground being a flat surface. 
In the next step, they use a monocular camera in combination with HOG~\cite{dalal2005histograms} and SVM~\cite{suykens1999least} algorithms to detect pedestrians and vehicles, while limiting the search area to bounding boxes of detections previously obtained from the LIDAR.
Einhorn~\etal~\cite{einhorn2011attention} use frontally mounted monocular camera and Extended Kalman Filters to reconstruct the scene and consequently detect potential obstacles. They propose attention-driven method for image areas where the obstacle situation is unclear and a more detailed scene reconstruction is necessary. \'{C}esi\'{c}~\etal~\cite{cesic2016radar} fuse detections from stereo cameras and RA\-DAR. Visually detected obstacle features are matched using stereo block matching and optical flow procedure. Filtered feature correspondences are projected to radar plane and passed to the multi-target tracking algorithm. Asvadi~\etal~\cite{asvadi20163d} propose obstacle detection method using LIDAR fused with measurements from inertial navigation system (GPS/IMU). The method mainly improves detection of pedestrians and cyclists. Li~\etal~\cite{li2016road} presented a method for road detection with image segmentation based on dark channel prior \cite{he2011single} and horizon estimation.

The majority of approaches that were developed for autonomous ground vehicles rely on estimation of the ground plane,
and cannot be readily applied to the aquatic environment of the USVs. A common practice for obstacle detection in marine environment is the use of RADAR, sonar, or LIDAR.
Almeida \etal~\cite{Almeida2009} proposed obstacle detection using on-board radar. They experienced difficulties in detecting small obstacles that were located in the near proximity (closer than \SI{200}{\meter}) of the boat.
The size and power consumption of radar units represent an additional challenge in application on small-sized USVs. Several approaches have therefore focused on obstacle detection using cameras.

Larson~\etal~\cite{larson2007advances} present advances in obstacle avoidance for USVs and point out the use of cameras. Their approach to obstacle detection with monocular camera relies on horizon estimation and image segmentation. On open sea, they use simple trigonometric calculations to estimate the horizon, while in marine environment and near shoreline, they use nautical charts, which provide them with analogous baseline distance to the shore. Guo~\etal~\cite{Guo2011} use an omni-directional camera to detect obstacles in water, based on the difference between two consecutive frames and foreground extraction. However, this method does not consider the dynamic nature of 
clouds and sky, nor their effect on the visual properties of water. This results in spurious difference between consecutive images, and thus incorrect foreground extraction due to weather conditions. 
Gal~\cite{Gal2011} uses edge detection approaches to detect the horizon line, and then search for obstacles 
below the estimated horizon. A major drawback of the method described in~\cite{Gal2011} is approximation of the sea-edge with a straight line. This assumption is often violated, especially in coastal waters and in 
marina, where the horizon does not correspond to the edge of water. The method described in~\cite{Gal2011} also relies on a sharp boundary between sea and sky when estimating the horizon line. In practice, however, 
this boundary is often blurred due to unfavorable weather conditions (haze, overwhelming cloudiness, fog), sun glitter, and reflections of the surrounding environment in water, making estimation of the exact position of the horizon difficult. Osborne~\etal~\cite{osborne2015temporally} propose a method capable of tracking objects on both visible and thermal imagery. They detect obstacles based on
agglomerative clustering of temporally stable features and track stable object clusters frame-to-frame. Cane~\etal~\cite{cane2016saliency} proposed obstacle detection and tracking based on saliency maps with wake and glint suppression and achieve excellent results on the open-seas dataset~\cite{andersson2016ipatch}, which is part of the PETS2016~\cite{patino2016pets} challenge.

Wang~\etal~\cite{Wang2011,wang2011real,Wang2012}
propose a vision-based obstacle detection that incorporates a stereo camera system. They use saliency detection in left camera to detect obstacles under the estimated sea edge and motion estimation to refine the output from saliency detection. Using epipolar constraints of stereo camera system and template matching, they search for correspondences of detected obstacles in right camera. In~\cite{Wang2012} they proposed disparity filtering to further improve position of the correspondences. However, their assumption of sharp boundary between sea and sky is often violated in practice. They also rely solely on the left camera for obstacle detection. Wang~\etal~\cite{wang2013stereovision} present stereo vision based obstacle detection. They estimate the sea surface by fitting a plane to the 3-D point set of the reconstructed scene. By aligning the reconstructed 3-D points with the sea surface, they compute occupancy grid and height grid, based on which the obstacles above water are detected. A calm sea may lack texture, which results in poor 3-D reconstruction of the sea, and consequently inaccurate sea surface estimation. In addition, such approach requires objects to significantly protrude through the water surface in order to  be distinguished as obstacles. This assumption is violated for small buoys and floating debris. Huntsberger~\etal~\cite{ROB:ROB20380} propose a stereo vision based navigation using a Hammerhead system. They generate dense range images, project range data into 2-D grid map, and perform spatial and temporal filtering on the map. For each map cell they compute a hazard probability, which serves for detecting and tracking discrete objects. However, the Hammerhead system cannot be used on small-sized USVs due to its dimensions, and suffers from the assumption that all obstacles stick out of the water.

Recently, Kristan~\etal~\cite{KristanCYB2015} proposed a graphical model for monocular obstacle detection via semantic segmentation of the observed marine scene. This state-of-the-art algorithm is most closely related to our work. The algorithm generates a water segmentation mask, and treats all objects in the water region as obstacles.
The model assumes that an image of marine environment can be partitioned into three distinct and approximately parallel semantic regions: sky at the top, ground or haze in the middle, and water at the bottom of the image. The semantic structure is enforced by fitting vertically distributed Gaussian components and regularizing the result with a Markov random field. This approach significantly outperforms the related approaches on the task of marine obstacle detection. It successfully detects obstacles protruding through the surface and floating obstacles, does not assume a straight water edge, and runs in real-time. Nevertheless, the SSM~\cite{KristanCYB2015} still fails in the presence of visual ambiguities. For example, when the boat faces open water and the horizon is hidden by the haze, the SSM approach drastically over- or under-estimates the extent of the water region.

%%%%%%%%%%%%%%%%%%%%%%%%%%%%%%%
% Implemented method overview %
%%%%%%%%%%%%%%%%%%%%%%%%%%%%%%%
\section{Augmented semantic segmentation model}\label{sec:algDesc}
\begin{figure}
	\centering
	\includegraphics[width=1\columnwidth]{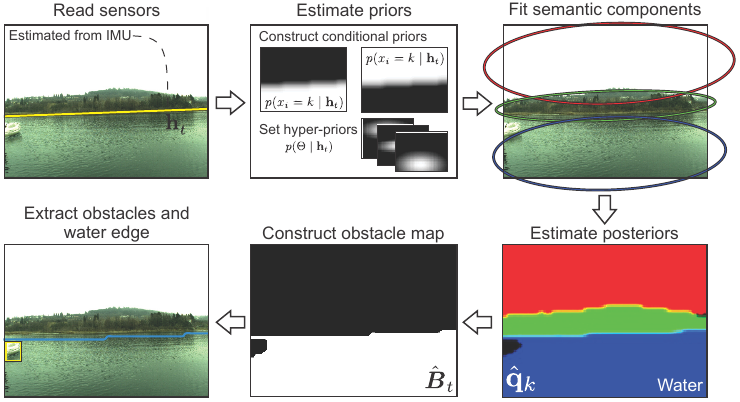}
	\caption{Outline of the ISSM. The IMU readings continuously modify hyper-priors of the semantic segmentation model, reducing the visual ambiguity and leading to robust edge of water estimation and improved obstacle detection.}%
	\label{fig:our_approach}
\end{figure}

Following the notation from \cite{KristanCYB2015}, the input image is represented by an array of values $\boldsymbol{Y} = \left\{ \boldsymbol{y}_i \right\}_{i=1:M}$, where \\
$\boldsymbol{y}_i = \left[ u_i, v_i, r_{i}, g_{i}, b_{i} \right]^T$ is a feature vector consisting of the pixel's position, $\left[ u_i, v_i \right]^T$, and RGB color values,
$\left[ r_{i}, g_{i}, b_{i} \right]$. In the segmentation model, each pixel is described with a four-component mixture model (\cref{fig:our_approach}),
\begin{align}
p(\boldsymbol{y}_i \mid \Theta, \boldsymbol{h}_t, \boldsymbol{\pi} ) = \sum\limits_{k=1}^3 \phi(\boldsymbol{y}_i \mid \mu_k, \Sigma_k) \tilde{\pi}_{ik} + \mathcal{U}(\boldsymbol{y}_i)\tilde{\pi}_{i4}, \label{eq:measurementmodel}
\end{align}
where $\phi(\cdot \mid \mu, \Sigma)$ are the Gaussians corresponding to the three main semantic regions, $\boldsymbol{h}_t$ denotes horizon line parameters at time $t$ and $\mathcal{U}(\cdot)$ is an additional uniform component that models the outliers.
In contrast to~\cite{KristanCYB2015} we make the per-pixel class priors, $\tilde{\pi}_{ik}$, explicitly depend on the prior probability of their labels, which are estimated from the
currently-available horizon parameters $\boldsymbol{h}_t$, i.e.,
\begin{equation}
  \tilde{\pi}_{ik} = \pi_{ik} p(x_i=k \mid \boldsymbol{h}_t), \label{eq:pixelpriors}
\end{equation}
where $\pi_{ik} = p(x_i = k)$ and $p(x_i=k \mid \boldsymbol{h}_t)$ is the prior computed from the current horizon position, as described in \cref{sec:maskIMU}.

To encourage segmentation into three approximately ver\-ti\-cally-distributed semantic structures, Kristan~\etal~\cite{KristanCYB2015} define hyper-priors $\varphi_0$ over the
Gaussian means in \eqref{eq:measurementmodel}. Since positions and orientations of Gaussians depend on roll and pitch of the boat, and thus on parameters of horizon line
in the image, we make this relation explicit, by defining the hyper-priors on the Gaussian parameters as
\begin{equation}
p(\Theta \mid \varphi_0, \boldsymbol{h}_t) = \prod\nolimits_{k=1}^3 \phi\left(\mu_k \mid \mu_{\mu_k}(\boldsymbol{h}_t), \Sigma_{\mu_k}(\boldsymbol{h}_t)\right). \label{eq:hyperpriors}
\end{equation}
The modified hyper-priors are defined as $\boldsymbol{\varphi}_0(\boldsymbol{h}_t) = \left\lbrace \boldsymbol{\mu}_{\mu_k}(\boldsymbol{h}_t), \boldsymbol{\Sigma}_{\mu_k}(\boldsymbol{h}_t) \right\rbrace_{k=1:3}$, where $\boldsymbol{\mu}_{\mu_k}(\boldsymbol{h}_t)$ and $\boldsymbol{\Sigma}_{\mu_k}(\boldsymbol{h}_t)$ are computed as described in \cref{sec:hyperPriors}.

The posteriors over pixel class labels are defined according to~\eqref{eq:measurementmodel} as $p_{ik}=\phi(\boldsymbol{y}_i \mid \mu_k, \Sigma_k) \tilde{\pi}_{ik}$ for $k\in [1,3]$ and $p_{i4}=\mathcal{U}(\boldsymbol{y}_i)\tilde{\pi}_{i4}$. Following Diplaros~\etal~\cite{Diplaros}, a smooth segmentation is encouraged by treating both the priors, $\boldsymbol{\pi} = \left\lbrace \pi_i \right\rbrace_{i=1:M}$, and the posteriors, $\boldsymbol{P} = \left\lbrace \boldsymbol{p}_i \right\rbrace_{i=1:M}$, over the pixel class labels as random variables, which form a Markov random field. The joint distribution over priors is approximated by $p(\boldsymbol{\pi}) \approx \prod\nolimits_i p(\pi_i \mid \pi_{N_i})$, where $\pi_{N_i}$ is a mixture distribution over the priors of the $i$-th pixel's neighbors. The potentials in the MRF corresponding to priors are 
\begin{equation}
p(\pi_i \mid \pi_{N_i}) \propto \exp{(-\frac{1}{2} E(\pi_i, \pi_{N_i}) )} ,
\end{equation}
with exponent defined as $E( \pi_i, \pi_{N_i}) = D(\pi_i \parallel \pi_{N_i}) + H(\pi_i)$. The term $D(\pi_i \parallel \pi_{N_i})$ is the Kullback-Leibler divergence, while $H(\pi_i)$ is the entropy. This term penalizes differences over neighboring priors and discourages uniform distributions in the priors. The distribution over posteriors, $p(\boldsymbol{P})$, is defined in the same way, leading to the following joint probability density function
\begin{align}\label{eq:cost1}
p(\boldsymbol{P}, \boldsymbol{Y}, \Theta, \boldsymbol{\pi} \mid \varphi_0) \propto  \exp \left[ \sum\limits_{i=1}^M \log p(\boldsymbol{y}_i , \Theta \mid \varphi_0) \right. \nonumber \\
\left. -  \frac{1}{2}\left( E(\pi_i, \pi_{N_i}) + E(\boldsymbol{p}_i,\boldsymbol{p}_{N_i}) \right) \right],
\end{align}
where $p(\boldsymbol{y}_i, \boldsymbol{\Theta} \mid \boldsymbol{\varphi}_0)$ is calculated by Bayes rule from \eqref{eq:measurementmodel} and \eqref{eq:hyperpriors}. Following the derivations in \cite{Diplaros} and \cite{KristanCYB2015}, the maximization of \eqref{eq:cost1} can be achieved by introducing auxiliary variables for the priors and posteriors, $\boldsymbol{s}_i$ and $\boldsymbol{q}_i$, leading to an EM-like algorithm with the E-step
\begin{align}
\hat{\boldsymbol{s}}_{{\cdot k}} &= (\xi_{s \cdot} \circ \boldsymbol{\pi}_{\cdot k} \circ (\boldsymbol{\pi}_{\cdot k} * \lambda))*\lambda_1, \nonumber\\
\hat{\boldsymbol{q}}_{{\cdot k}} &= (\xi_{q \cdot} \circ \boldsymbol{p}_{\cdot k} \circ (\boldsymbol{p}_{\cdot k} * \lambda))*\lambda_1, \nonumber\\
\pi^\mathrm{opt}_{\cdot k} &= (\hat{\boldsymbol{s}}_{{\cdot k}} + \hat{\boldsymbol{q}}_{{\cdot k}})p(\boldsymbol{x}_{\cdot} = k \mid \boldsymbol{h}_t)/4, \label{eq:EStep}
\end{align}
where $\circ$ and $*$ denote the Hadamard product~\cite{horn1990hadamard} and convolution, respectively. $\lambda$ is a small discrete Gaussian kernel (e.g., $3 \times 3$) with its central element set to zero and its elements summing to one, while $\lambda_1 = 1 + \lambda$. The main difference between \eqref{eq:EStep} and the equivalent step in~\cite{KristanCYB2015} is the addition of the prior probability of a pixel belonging to the water class, $p(x_{\cdot} = k \mid \boldsymbol{h}_t)$, conditioned on the horizon-line position, $\boldsymbol{h}_t$ (we describe this prior in \cref{sec:maskIMU}).

\begin{figure}
	\centering
	\includegraphics[width=1\columnwidth]{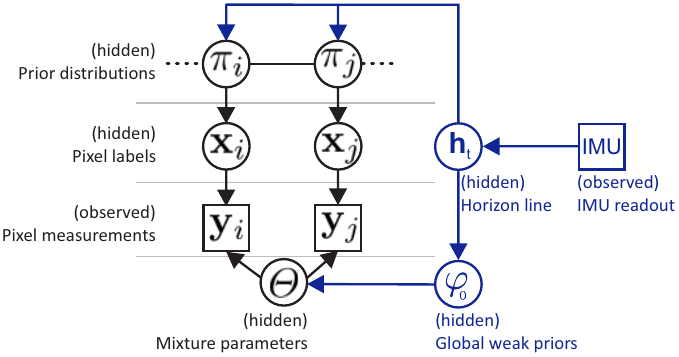}
	\caption{The proposed graphical model. The extension from the original model~\cite{KristanCYB2015} is shown in blue.}
	\label{fig:graphical_model}
\end{figure}

The M-step recomputes means and variances of the Gaussians in \eqref{eq:measurementmodel} using the horizon-dependent hyper-priors as
\begin{align}
    \mu_k^\mathrm{opt} &= \beta_k^{-1} \left[ \Lambda_k \left( \sum\limits_{i=1}^M \hat{q}_{ik} \boldsymbol{y}_i^T \right) \Sigma_k^{-1} - \mu_{\mu_k}^T \left( \boldsymbol{h}_t \right) \Sigma_{\mu_k}^{-1} \right]^T, \nonumber\\
    \Sigma_k^\mathrm{opt} &= \beta_k^{-1} \sum\limits_{i=1}^M \hat{q}_{ik}(\boldsymbol{y}_i - \mu_k)(\boldsymbol{y}_i - \mu_k)^T, \label{eq:MStep}
\end{align}
where $\beta_k=\sum\nolimits_{i=1}^M \hat{q}_{ik}$ and $\Lambda_k=(\Sigma_k^{-1} + \Sigma_{\mu_k}^{-1})^{-1}$. For clearer exposition, we illustrate the differences between the original~\cite{KristanCYB2015} and the proposed graphical model in \cref{fig:graphical_model}. In practice, the alternating computations of EM steps \eqref{eq:EStep} and \eqref{eq:MStep} require only few iterations to converge.

\subsection{Estimation of conditional prior distributions}
\label{sec:maskIMU}
Our semantic segmentation model (\cref{fig:graphical_model}) utilizes the IMU information in per-pixel class priors through the conditional prior distributions $p(x_i = k \mid \boldsymbol{h}_t)$ \eqref{eq:pixelpriors}, which are computed from the horizon location.

\begin{figure}
	\centering
	\includegraphics[width=1\columnwidth]{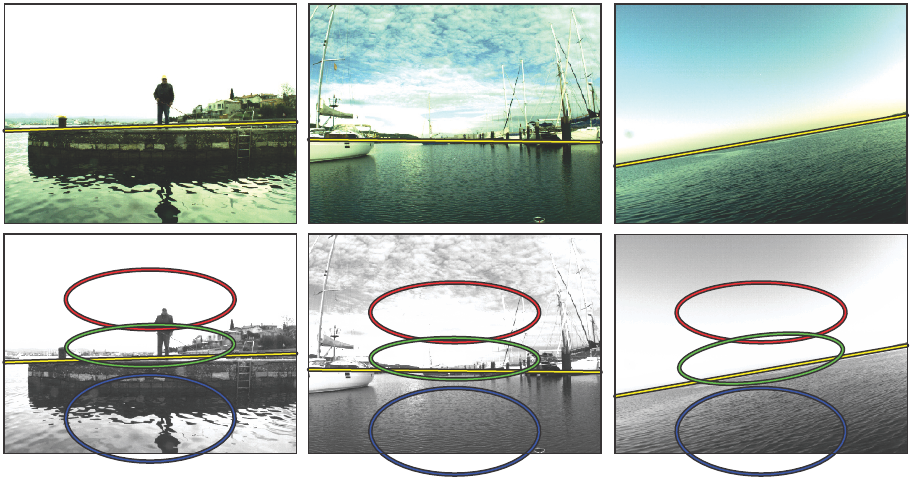}
	\caption{The water edge does not always match the horizon (yellow line). The bottom row shows hyper-priors of semantic component centers, modulated by the estimated position and angle of the horizon.}
  \label{fig:horizonLinePriors}
\end{figure}

When the boat is facing open water, the horizon tends to approximately match the water edge. But this is not true when the boat is close to the shore and partially facing inland. In these situations, a part of the region below the horizon may instead correspond to the land (see \cref{fig:horizonLinePriors}). 
Nevertheless, the water is never located above the horizon and the sky is never located below the horizon. These observations are used in our definition of the conditional prior distributions.

\begin{figure}
	\centering
	\includegraphics[width=1\columnwidth]{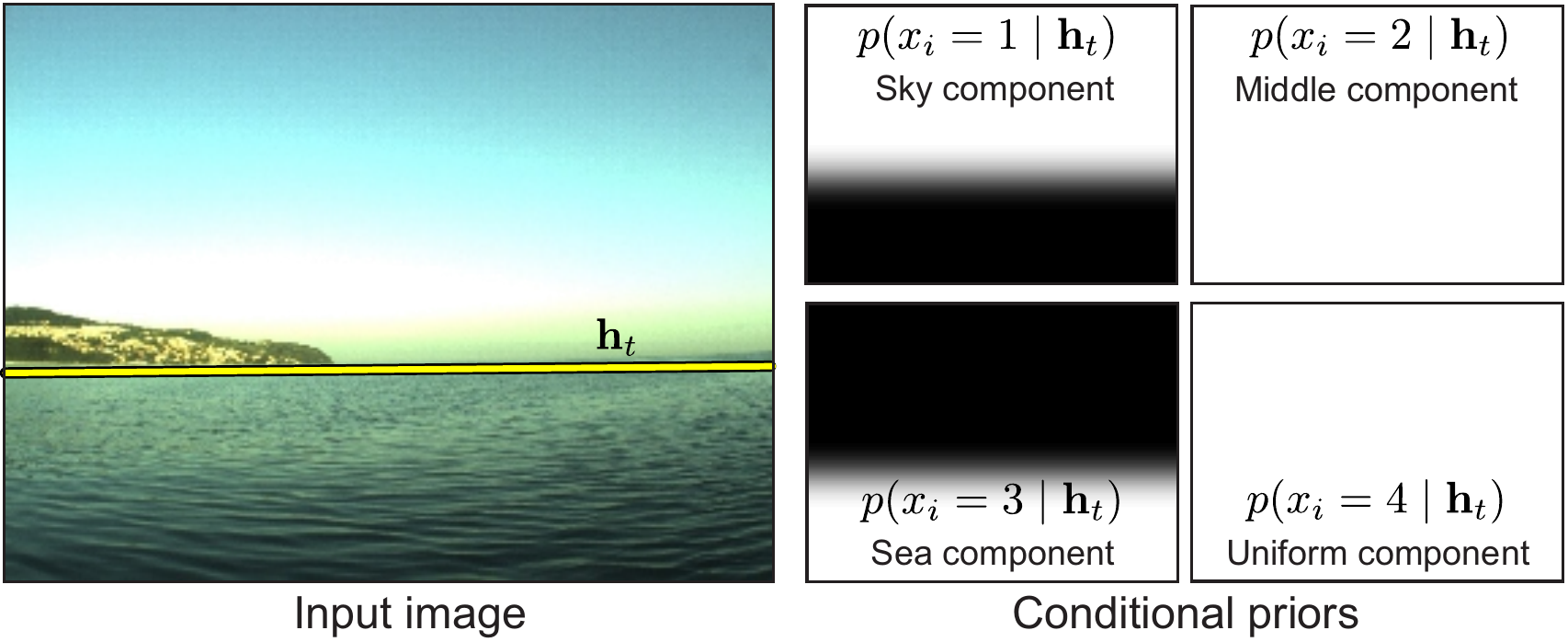}
	\caption{An input image and the corresponding conditional priors $p(x_i = k \mid \boldsymbol{h}_t)$ of semantic components. The horizon line $\boldsymbol{h}_t$, estimated from the IMU readings, is shown as yellow line.}%
	\label{fig:modification_priors}
\end{figure}

The conditional prior $p(x_i = k \mid \boldsymbol{h}_t)$ of the water component is set to zero for pixels above the horizon and to one elsewhere. Similarly, the conditional prior of the sky component is set to zero for pixels below the horizon and to one elsewhere. 
The conditional priors of the remaining two components are set to uninformative priors, i.e., all values are set to one. To account for the noise in horizon estimates, for example due to synchronization issues or nonrigid coupling between the camera and IMU, 
all conditional priors are blurred by a Gaussian with a small variance. An example of the estimated conditional priors is shown in \cref{fig:modification_priors}.

\subsection{Estimation of the Gaussian hyper-priors} 
\label{sec:hyperPriors}
Because the three Gaussian components in \eqref{eq:measurementmodel} correspond to semantic elements in the scene, their spatial parts change with changing roll and pitch of the boat. For example, upward pitch moves the horizon in the image downwards, which should affect the likely position of the Gaussian components. The prior on their position is captured by the hyper-priors $\boldsymbol{\varphi}_0$ in \eqref{eq:hyperpriors}.

The original SSM~\cite{KristanCYB2015} applies fixed hyper-priors (\ref{eq:hyperpriors}), which are learned from training images with manual annotations of the three semantic regions. We follow a similar
procedure to learn horizon-dependent hyper-priors, $\boldsymbol{\varphi}_0(\boldsymbol{h}_t) = \left\lbrace \boldsymbol{\mu}_{\mu_k}(\boldsymbol{h}_t), \boldsymbol{\Sigma}_{\mu_k}(\boldsymbol{h}_t) \right\rbrace$,
that change with USV motion. We annotated the horizon location in each of the training images from SSM~\cite{KristanCYB2015}. For each semantic component, the average vertical displacement to the horizon line was computed over all images. Thus during the USV operation, the vertical position of the learned mean priors $\boldsymbol{\mu}_{\mu_k}(\boldsymbol{h}_t)$ are set to the learned displacement from the current horizon estimate. \cref{fig:horizonLinePriors} shows examples of our horizon-based dynamic Gaussian priors adjustment.

Since the middle component usually moves with the horizon, we modify the corresponding hyper-prior of the middle Gaussian component as well. In particular, the spatial covariance of the middle component, $\Sigma_{\mu_2}$, in 
(\ref{eq:hyperpriors}) is contained in the upper-left $2 \times 2$ sub-matrix $\Sigma_{\mu_{2{\mathrm{pos}}}}$. Let $\boldsymbol{R}_{\boldsymbol{h}_t}$ be a $2 \times 2$ rotation matrix that rotates by the angle $\gamma_{\boldsymbol{h}_t}$. The hyper-prior covariance of the middle component is obtained by replacing its spatial part with the following proximal projection with the angle equal to $\gamma_{\boldsymbol{h}_t}$
\begin{equation}
\tilde{\Sigma}_{2{\mathrm{pos}}} = \boldsymbol{R}_{\boldsymbol{h}_t}^T\left(\left(\boldsymbol{R}_{\boldsymbol{h}_t}\Sigma_{2{\mathrm{pos}}}\boldsymbol{R}_{\boldsymbol{h}_t}^T\right) \circ \boldsymbol{I}_{2}\right)\boldsymbol{R}_{\boldsymbol{h}_t},
\label{eq:modHyperCov}
\end{equation}
where $\boldsymbol{I}_{2}$ is a $2 \times 2$ identity matrix. %and $\circ$ is a Hadamard product~\cite{horn1990hadamard}. 
The angle $\gamma_{\boldsymbol{h}_t}$ corresponds to the slope of the projected horizon line $\boldsymbol{h}_t$ in the image.

\subsection{Estimation of horizon line from IMU measurements}
\label{sec:horest}
By definition, the horizon line in the image corresponds to projection of points that lie infinitely far away on the water surface. Since all parallel planes project to the same horizon line~\cite{MultipleViewGeometryZisserman}, the projection is independent from the camera height, and is governed only by the camera-to-plane rotation.

Let $\boldsymbol{X}^\mathrm{usv}$ denote 3-D coordinates of a point, located on the water surface far 
away in front of the camera, and expressed in the coordinate system of the USV. Let $\boldsymbol{R}_\mathrm{cam}^\mathrm{usv}$ denote the matrix describing the rotation between coordinate systems of the USV and the camera. 
The point $\boldsymbol{X}_i^\mathrm{usv}$ is projected into camera's image plane according to
\begin{equation}
\lambda_{\mathrm{c}}\boldsymbol{x}_i = \boldsymbol{K}\boldsymbol{R}_\mathrm{cam}^\mathrm{usv}\boldsymbol{X}_i^\mathrm{usv}, \label{eq:projtocam}
\end{equation} 
where $\boldsymbol{K}$ is the camera calibration matrix, which is estimated during the calibration process.

In our approach, the points $\boldsymbol{X}_i^{\mathrm{usv}}$ forming the horizon are obtained from the IMU measurements. Let $\boldsymbol{R}_\mathrm{usv}^\mathrm{imu}$ denote the rotation of IMU
relative to the USV, and let $\boldsymbol{R}_\mathrm{imu}$ denote the rotation of IMU relative to the water surface (i.e., the readout of Euler angles from the IMU).
We make a reasonable assumption that the Z-axes of IMU and camera are approximately aligned by design. In principle, these geometric relations are sufficient to calculate the vanishing points, which can be directly used to estimate the horizon line. 
However, as experimentally demonstrated in \cref{sec:parameters}, projection of vanishing points into input image tends to incur inaccuracies. This is because vanishing points likely project outside of the image boundaries, while the calibrated
radial distortion model reliably estimates distortion only for points inside the image.

Therefore, we instead obtain horizon line by projecting two points, $\{\boldsymbol{X}_1^\mathrm{imu}$ and $\boldsymbol{X}_2^\mathrm{imu} \}$ lying on the XZ-plane of the IMU coordinate system at horizontal angles $\pm \alpha_h$ (constrained by the camera FOV) and at finite distance $Z = l_{\mathrm{dist}}$. 
These points are rotated into a plane parallel to the water surface by relation
\begin{equation}
\boldsymbol{X}_i^\mathrm{usv} = \boldsymbol{R}_\mathrm{imu}\left(\boldsymbol{R}_\mathrm{usv}^\mathrm{imu}\right)^{-1}\boldsymbol{X}_i^\mathrm{imu}, \label{eq:projtousv} 
\end{equation}
where $\boldsymbol{X}_i^\mathrm{imu}$ and $\boldsymbol{X}_i^\mathrm{usv}$ denote a point before and after rotation, respectively. The rotated points $\{\boldsymbol{X}_1^\mathrm{usv}, \boldsymbol{X}_2^\mathrm{usv} \}$ are projected into the image using (\ref{eq:projtocam}) while taking into account radial distortion. The horizon is estimated by fitting a line to the projected radially distorted points.

\subsection{Obstacle detection by semantic segmentation} \label{sec:obst_detbyssm}
Obstacles are detected by post-processing the output of the semantic segmentation algorithm, where we follow the approach from~\cite{KristanCYB2015}. Fitting the semantic model by~\eqref{eq:EStep} and~\eqref{eq:MStep} to the input image results in pixel-wise a-posteriori probability distributions $\hat{\boldsymbol{q}}_{ik}$, which indicate the probability of each pixel belonging to one of the four semantic components (\cref{fig:our_approach}). Pixels are labeled as water if their a-posteriori probability is maximized for the water component, which in our case is indexed as $k=3$.

Let $\boldsymbol{B}_t$ be the water region mask and let $B_i \in [0, 1]$ indicate the water label of the $i$-th pixel. The water region mask is therefore constructed by
\begin{equation}
B_i = \begin{cases}
    1,	& \arg\max_k \hat{\boldsymbol{q}}_{ik} = 3\\
    0,  & \text{otherwise}
\end{cases}.
\label{eq:water_region_image}
\end{equation}

From water region mask $\boldsymbol{B}_t$ we extract the largest connected region, resulting in an obstacle map $\hat{\boldsymbol{B}}_t$. The blobs of non-water pixels surrounded by water pixels in $\hat{\boldsymbol{B}}_t$ correspond to potential obstacles. Non-maximum suppression is performed on $\hat{\boldsymbol{B}}_t$ to merge nearby blobs in order to avoid multiple detections of the same obstacle. The water edge is defined as the largest connected outer edge of regions in the obstacle map $\hat{\boldsymbol{B}}_t$. The proposed obstacle detection algorithm ISSM is summarized by 
\cref{alg:obstacle_detection_mono,fig:our_approach}.

% Algoritem 1 begin
\begin{algorithm}
  %\scriptsize
  \caption{Monocular obstacle detection algorithm ISSM}
  \label{alg:obstacle_detection_mono}
  \begin{algorithmic}[1]
    \Require 
    \Statex Pixel features $\boldsymbol{Y} = {\left\lbrace \boldsymbol{y}_i \right\rbrace}_{i=1:M}$, IMU read-out, model from previous time-step $\boldsymbol{\Theta}_{t-1}$ and $\hat{\boldsymbol{q}}_{t-1}$.
    
    \Ensure 
    \Statex Obstacle image map $\hat{\boldsymbol{B}}_t$, water edge $\boldsymbol{e}_t$, detected objects ${\left\lbrace \boldsymbol{o}_i \right\rbrace}_{i=1:N_{obj}}$ and model parameters $\boldsymbol{\Theta}_t$, $\hat{\boldsymbol{q}}_t$.
    \State Compute $\boldsymbol{h}_t$ from IMU (\cref{sec:horest}).
    \State Estimate conditional prior distributions (\cref{sec:maskIMU}) and Gaussian hyper-priors (\cref{sec:hyperPriors}) using $\boldsymbol{h}_t$.
    \State Estimate the per-pixel posteriors by \eqref{eq:EStep}, \eqref{eq:MStep} and \eqref{eq:modHyperCov}.%\eqref{5,6,9}.
    \State Calculate obstacle image map $\hat{\boldsymbol{B}}_t$, water edge $\boldsymbol{e}_t$ and obstacles below water edge ${\left\lbrace \boldsymbol{o}_i \right\rbrace}_{i=1:N_{obj}}$ (\cref{sec:obst_detbyssm}).
  \end{algorithmic}
\end{algorithm}
% Algoritem 1 end

\section{Stereo verification for improved detection} \label{sec:stereoDetection}
The ISSM from \cref{alg:obstacle_detection_mono} generates a list of potential obstacles based on their distinctiveness from the water. The list may contain false positives due to water droplets and reflections of sun or nearby environment. In addition, small objects are more likely to be incorrectly segmented, resulting in false negative (missing) detections. 
To address both of these issues, we separately apply the ISSM to left and right camera of the USV's on-board stereo system, and use epipolar constraints and template matching to consolidate the obtained lists of detected obstacles.

\begin{figure}
	\centering
	\includegraphics[width=1\columnwidth]{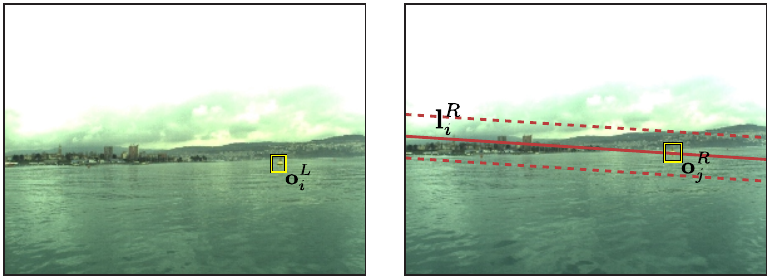}
	\caption{Obstacle detections $\boldsymbol{o}_i^L$ and $\boldsymbol{o}_j^R$ in left camera and right camera (yellow bounding box). The red line denotes the epipolar line $\boldsymbol{l}_i^R$ corresponding to the obstacle $\boldsymbol{o}_i^L$, while the dashed line denotes the extended search region.}%
	\label{fig:epipolar_constraints}
\end{figure}

Let $\{ \boldsymbol{o}^L_i \}_{i=1:N_L}$ and $\{ \boldsymbol{o}^R_j \}_{j=1:N_R}$ be the list of tentative detected obstacles, obtained by ISSM in left and right camera, respectively. Each detection is parameterized by its position and size in the image: $\boldsymbol{o}_i=[ u_i, v_i, w_i, h_i]^T$. According to the epipolar constraint, an object detected in the left image should have a corresponding detection along the epipolar line in the right image. Due to imperfect segmentation and calibration, the correspondence center will not lie exactly on the epipolar line, but within its vicinity, which is set to the obstacle bounding box diagonal in our application (\cref{fig:epipolar_constraints}).

Since multiple detections might agree with the epipolar constraint, a further verification is required. A template of the detected obstacle in the left image is extracted and matched against the potential candidates in the right image (and vice versa) by means of normalized cross correlation (NCC)~\cite{lewis1995fast}. 
In some cases segmentation might successfully segment only part of the object. Therefore matching is performed in a search region equal to the object size enlarged by a factor $\theta_\mathrm{S1}$.

In particular, an NCC response map is computed separately between the template and search region on each color channel and the result is averaged over the three channels yielding a single response map. A match is accepted if the maximum of the response map exceeds a pre-defined threshold $\theta_\mathrm{ncc}$.

The epipolar verification addresses the problem of false positives. However, some obstacles may be detected only in one image and not in the other. Therefore, all tentative detections that have not been successfully matched via the epipolar verification are revisited. For each such detection $\boldsymbol{o}^L_i$ in the left image, NCC verification is performed in a region in the right image. In absence of a detected candidate in the right image, the region is centered 
at the same coordinates as $\boldsymbol{o}^L_i$ and the region size is set to $\boldsymbol{o}^L_i$ enlarged by a factor $\theta_\mathrm{S2}$. The detection is accepted if maximum response of the NCC exceeds pre-defined threshold $\theta_\mathrm{ncc}$. The same approach is applied to the remaining tentative detections from the right camera. All tentative obstacles that do not pass this verification are discarded. The ISSM with added stereo verification is denoted as ISSM\textsubscript{S} in the remainder of the paper, and is summarized in \cref{alg:obstacle_detection_stereo}.

% Algoritem 2 begin - obstacle verification
\begin{algorithm}
  %\scriptsize
  \caption{Stereo obstacle detection algorithm ISSM\textsubscript{S}}
  \label{alg:obstacle_detection_stereo}
  \begin{algorithmic}[1]
    \Require 
    \Statex Pixel features $\boldsymbol{Y} = {\left\lbrace \boldsymbol{y}_i \right\rbrace}_{i=1:M}$, horizon line parameters $\boldsymbol{h}_t$, model from previous time-step $\boldsymbol{\Theta}_{t-1}$ and $\hat{\boldsymbol{q}}_{t-1}$.
    
    \Ensure
    \Statex Obstacle image map $\hat{\boldsymbol{B}}_t$, Water edge $\boldsymbol{e}_t$, detected objects ${\left\lbrace \boldsymbol{o}_i \right\rbrace}_{i=1:N_{obj}}$, model parameters $\boldsymbol{\Theta}_t$ and $\hat{\boldsymbol{q}}_t$.
		\State Obtain obstacles $\{ \boldsymbol{o^L}_i \}_{i=1:N_{obj_L}}$ and $\{ \boldsymbol{o^R}_j \}_{j=1:N_{obj_R}}$ by independently applying \cref{alg:obstacle_detection_mono} in left and right camera.
        \State Match obstacles from left and right camera using epipolar constraints and NCC verification. \label{alg:matching_first_pass}
        \State Perform a brute-force template matching for all unpaired detections.
  \end{algorithmic}
\end{algorithm}
% Algoritem 2 end

\section{The camera-IMU calibration} 
\label{sec:cameraImuCalibration}
Projection of horizon from IMU into camera (\ref{eq:projtocam}, \ref{eq:projtousv}) depends on three rotation matrices: $\boldsymbol{R}_{\mathrm{imu}}$, $\boldsymbol{R}_{\mathrm{cam}}^{\mathrm{usv}}$ and $\boldsymbol{R}_{\mathrm{usv}}^{\mathrm{imu}}$. The $\boldsymbol{R}_\mathrm{imu}$ relates the IMU 
coordinate system to the world coordinate system and is constructed from the IMU readings at each time-step. The $\boldsymbol{R}_\mathrm{cam}^\mathrm{usv}$
and $\boldsymbol{R}_\mathrm{usv}^\mathrm{imu}$ relate the camera and IMU coordinate systems to the USV coordinate system, respectively, and have 
to be estimated by calibration. We devised a practical approach for such calibration, which we outline in this section.

\begin{figure}
  \centering
  \includegraphics[width=1\columnwidth]{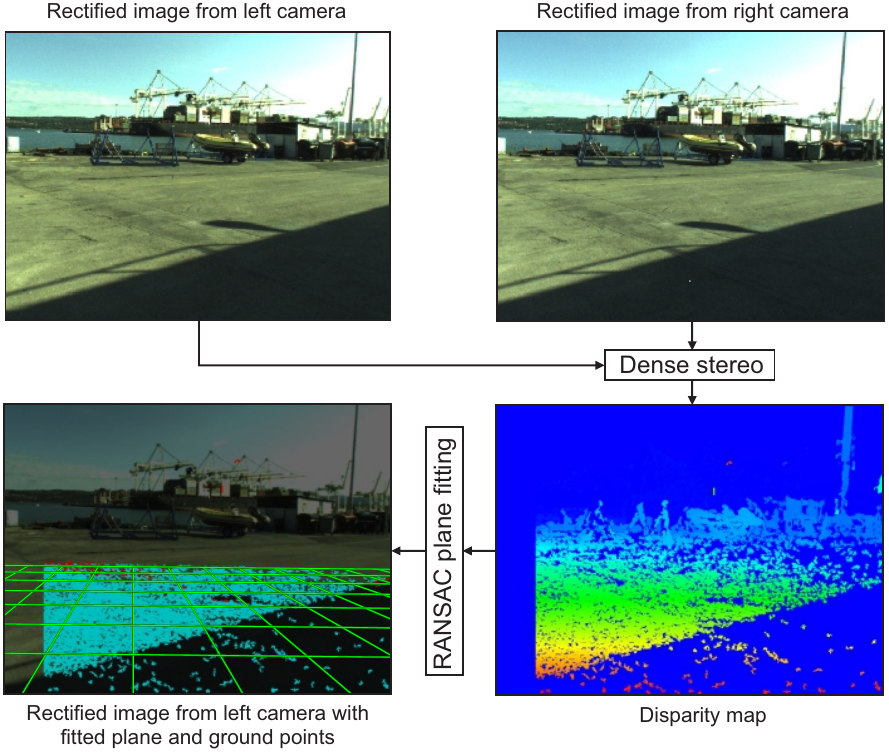}
  \caption{Plane fitting estimation for camera-IMU calibration. Points which meet the threshold $\theta_{\mathrm{dist}}$ condition are shown together with grid of the fitted plane in the bottom-left image.}
  \label{fig:plane_fitting}
\end{figure}

During calibration, the USV is placed on a solid flat ground, for example a parking lot in the marina (\cref{fig:plane_fitting}). Since the world- and USV-coordinate systems are aligned on
a flat surface, the $\boldsymbol{R}_\mathrm{usv}^\mathrm{imu}$ is directly obtained from the IMU readout. Estimation of the $\boldsymbol{R}_\mathrm{cam}^\mathrm{usv}$,
however, is not as straight forward, and requires computation of the floor-plane orientation with respect to camera. In principle, this could be achieved by placing 
a flat square marker on the floor in front of the camera, and estimating the rotation from the corresponding homography.

However, our USV is equipped with a stereo camera, which allows us to compute the point cloud, and use RANSAC \cite{ransac} to fit a ground plane to the points 
whose distance to boat does not exceed a predefined threshold $\theta_\mathrm{dist}$. The process is outlined in \cref{fig:plane_fitting}. The rotation matrix $\mathbf{R}_\mathrm{cam}^\mathrm{usv}$ is then computed from the normal of the estimated ground plane. The normal provides only rotations around X- and Z-axes. Since the camera Z-axis is in the direction of heading, we set the rotation around Y-axis to zero.

%%%%%%%%%%%%%%%
%%% DATASET %%%
%%%%%%%%%%%%%%%
\section{Multimodal marine obstacle detection dataset}
\label{sec:dataset}
Due to lack of a realistic, publicly-available USV dataset comprising annotated stereo video and IMU data, we captured our own dataset, called Marine obstacle detection dataset 2 (Modd 2). It contains 28 video sequences of variable length, amounting to \SI{11675}{} frames at resolution of \SI{1278x958}{} pixels.
This is currently the biggest marine obstacle detection dataset to date and we plan to make it publicly available as a part of this paper.

\begin{figure}
	\centering
	\includegraphics[width=1\columnwidth]{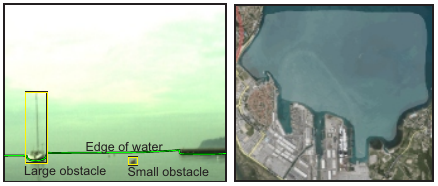}
	\caption{Left: An example of annotated frame. Right: The white highlighted area denotes the region in the coastal waters of Koper, Slovenia where the USV has acquired the data.}%
	\label{fig:AreaAndUSV}
\end{figure}

\begin{figure}
	\centering
	\includegraphics[width=1\columnwidth]{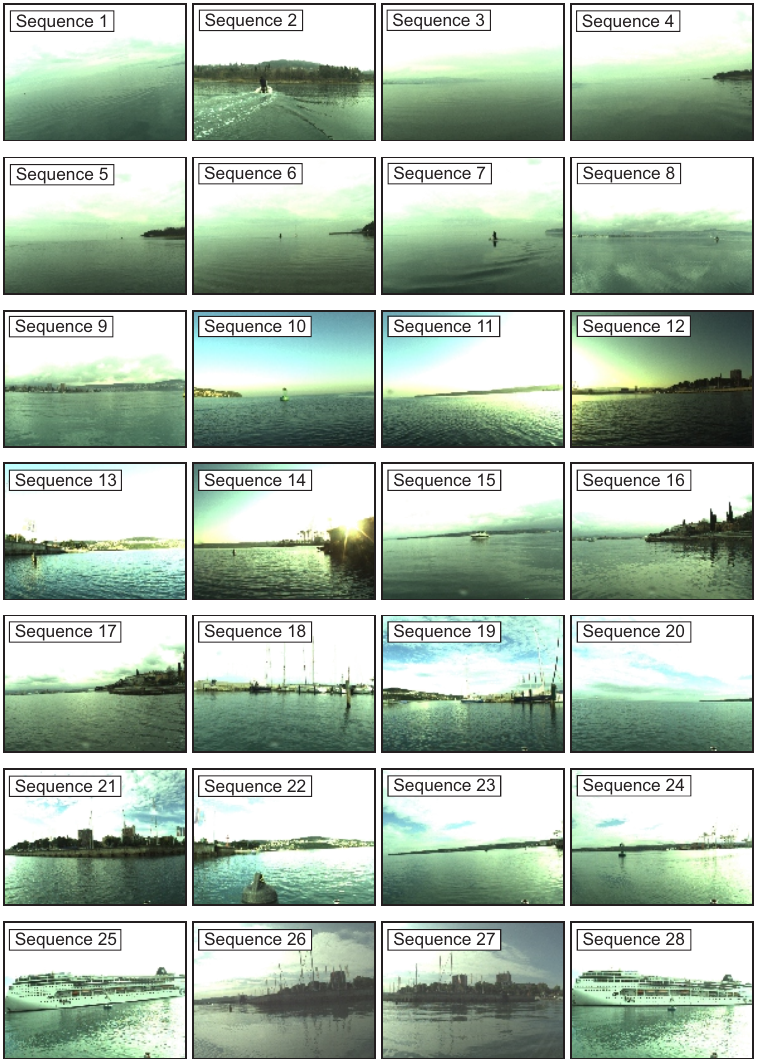}
	\caption{Random frame from each sequence in our dataset.}%
	\label{fig:dataset_images}
\end{figure}

The dataset was captured over a period of approximately $15$ months in the gulf of Koper,
Slovenia (area shown in~\cref{fig:AreaAndUSV}), using the USV developed by Harpha Sea, d.o.o. The USV uses a steerable thrust propeller with a small turn radius for guidance, and can reach the maximum velocity of \SI{2.5}{\meter\per\second}. The USV is equipped with a compass, IMU unit and a stereo system \emph{Vrmagic VRmMFC}, which consists of two \emph{Vrmagic VRmS-14/C-COB} CCD sensors, \emph{Thorlabs MVL4WA} lens with \SI{3.5}{\milli\meter} focal length, maximum aperture of $\mathrm{f}/1.4$, and a \SI{132.1}{\degree} FOV. The stereo system is mounted approximately \SI{0.7}{\meter} above the water surface, and faces forward. Both cameras are connected to the on-board computer through USB-2.0 bus and are thus capable of capturing the video of given resolution at $10$ frames per second. The cameras automatically 
adjust the aperture according to the lighting conditions. Examples of the sequences from the dataset are shown in~\cref{fig:dataset_images}.

\begin{figure}
\centering
  \includegraphics[width=1\columnwidth]{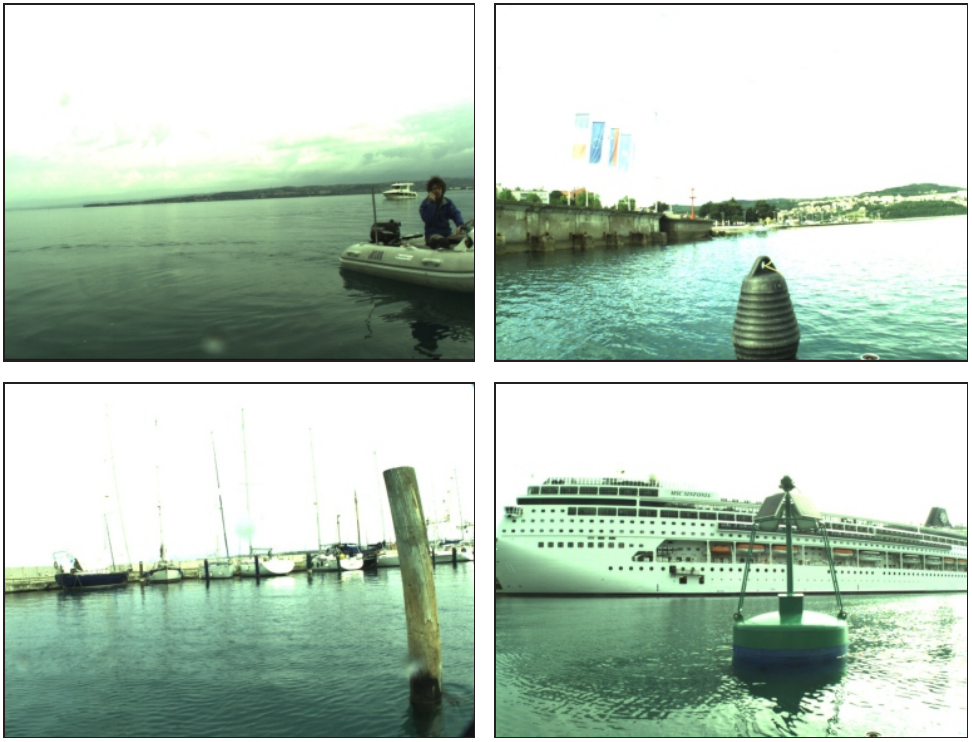}
  \caption{Left: Examples where the USV is sailing in the near proximity of the obstacles, passing them by. Right: Examples where the USV is sailing straight to the obstacle and eventually collides with it.}
  \label{fig:examplesNearProximityDirect}
\end{figure}

During the dataset acquisition, the USV was manually guided by an expert, who was instructed to simulate realistic navigation scenarios with situations in which an obstacle
may present a danger to the USV. Such situations include sailing straight in the direction of an obstacle (\cref{fig:examplesNearProximityDirect} left), and sailing in the near proximity of an obstacle (\cref{fig:examplesNearProximityDirect} right). The dataset was recorded over a period of several months at different times of day and under various weather conditions to maximize the visual diversity.

We followed the obstacle annotation protocol from \cite{KristanCYB2015}. Each frame was manually annotated by a human annotator and later verified
by an expert. The edge of water is annotated by a polygon, while obstacles are annotated with bounding boxes. The annotated obstacles are further divided into two
classes --- large obstacles (whose bounding box straddles the sea edge), and small obstacles (whose bounding box is fully located below the sea edge polygon). For completeness, the horizon is annotated on frames where the USV is facing away from the shore with the horizon clearly visible (see \cref{fig:AreaAndUSV}).

\begin{figure}
  \centering
  \includegraphics[width=1\columnwidth]{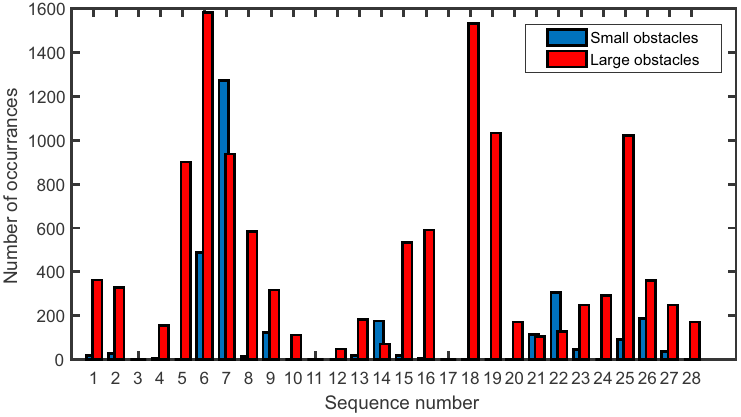}
  \caption{Number of obstacle occurrence per video sequence. Red color represents large obstacles and blue color represents small obstacles.}
  \label{fig:obstaclesPerSequence}
\end{figure}

\begin{figure}
  \centering
  \includegraphics[width=1\columnwidth]{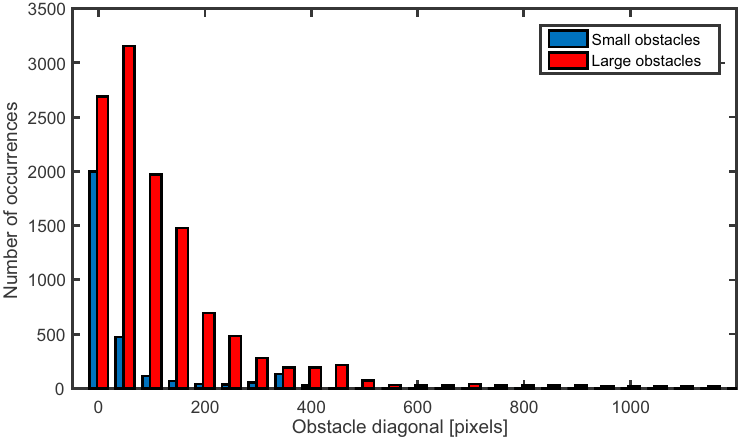}
  \caption{Distribution of obstacle sizes. Red color represents large obstacles and blue color represents small obstacles. Size is measured as diagonal length of obstacle bounding box in pixels.}
  \label{fig:distributionSizeObstacles}
\end{figure}

\cref{tab:datasetDetails} summarizes general properties of the dataset sequences and conditions that might affect the IMU accuracy and obstacle detection. Two sequences contain extreme sudden change of motion and boat rolling and pitching. Sun glitter is present in 12 sequences, while 16 sequences contain significant reflection of the environment in the water. Twenty-five out of 28 sequences contain at least one occurrence of an obstacle. The average number of obstacles per frame is $0.614$ with standard deviation $0.945$. The distribution of obstacle occurrence per video sequence is shown in \cref{fig:obstaclesPerSequence}, while \cref{fig:distributionSizeObstacles} shows the distribution of obstacle sizes.

\begin{table}[]
\centering
\caption{Statistics of Modd 2 video sequences. Number of frames in a sequence ($\mathrm{n_{frm}}$), average number of obstacles per frame in a sequence ($\mathrm{n_{obj}}$), informations of possible occurring extreme conditions (sudden movement of the USV, sun glitter and/or environment reflections in the water).}
\input{dataset_info}
\label{tab:datasetDetails}
\end{table}

\begin{figure}
	\centering
	\includegraphics[width=1\columnwidth]{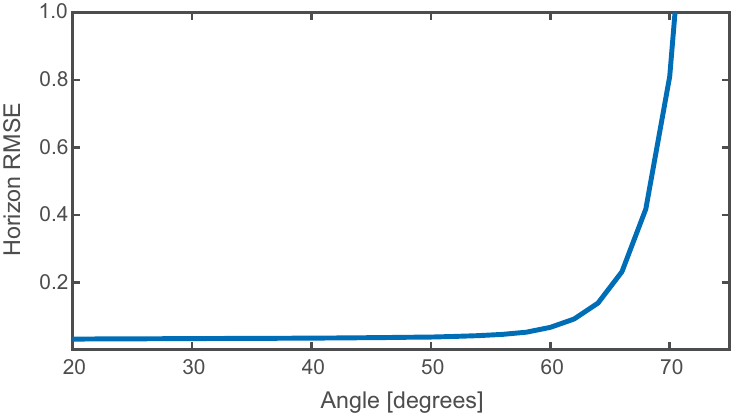}
	\caption{Accuracy of horizon line estimation with respect to the angle between points on the water surface. The RMSE of horizon is scaled to interval $\left[0, 1\right]$ to remove the effect of resolution.}%
	\label{fig:angles_graph}
\end{figure}

%%%%%%%%%%%%%%%%%%%%%%%%%%%%%%%
%%         Experiments       %%
%%%%%%%%%%%%%%%%%%%%%%%%%%%%%%%
\section{Experiments} \label{sec:experiments}
The prototype of our stereo IMU-assisted semantic segmentation algorithm (ISSM\textsubscript{S}) was implemented in Matlab. All experiments were performed on a desktop PC with the Intel Core i7-7700 \SI{3.6}{\giga\hertz} CPU. We plan to make the code publicly available as a part of this paper.

This section is structured as follows. \Cref{sec:parameters} provides relevant implementation details. \Cref{sec:mono_eval} compares our approach to the state-of-the-art on a monocular setup, while \cref{sec:ablationstudy} provides an ablation study. In \cref{sec:stereo_eval}, we compare our approach to the state-of-the-art in a stereo setup.
\Cref{sec:speed} analyzes the speed of our algorithm and its individual components. Lastly, \cref{sec:extreme_conditions} reports performance under extreme conditions, while the failure cases are discussed in \cref{sec:failure_cases}.

\subsection{Implementation details and parameters}
\label{sec:parameters}
As our approach is based on the state-of-the-art SSM~\cite{KristanCYB2015} algorithm, it uses the same parameters as SSM, listed in~\cite{KristanCYB2015}. The hyper-priors have been trained on the same sequences as~\cite{KristanCYB2015}, with additional horizon annotations (\cref{sec:dataset}). Segmentation is performed in the RGB colorspace.

Projection of the horizon into camera (\cref{sec:horest}) requires points that lie far away in front of the camera. Due to the Earth curvature it is sufficient to generate points at distance of $l_{\mathrm{dist}} = 3.57 \cdot 10^3\sqrt{h_{\mathrm{cam}}}$ m~\cite{horizonDistance}, where $h_{\mathrm{cam}}$ is the camera height in meters.

As noted in \cref{sec:horest}, the points should be within the camera FOV where the radial distortion model is most accurate. To demonstrate this, we have generated the points at various angles and measured the root-mean-square-error of the fitted horizon line. The results are shown in~\cref{fig:angles_graph}. Observe that the error is kept consistently low for a range of angles below \SI{56}{\degree} and exponentially increases for larger angles. In our experiments we thus set the angle to $\alpha_h = 40^{\circ}$.

The camera-IMU calibration (\cref{sec:cameraImuCalibration}) requires setting a threshold $\theta_{\mathrm{dist}}$, which specifies the region in front of the camera used to estimate the plane by RANSAC~\cite{ransac}. The calibration was performed in a marina parking lot (shown in \cref{fig:plane_fitting} top-left). The area of approximately ten meters in front of the USV contained no obstacles, thus we set the threshold to \SI{10}{\meter}. A point-to-plane distance threshold in RANSAC was set to $\SI{1}{\meter}$ and the maximum number of iterations was set to \SI{1000}{}. The number of maximum iterations was set to a standard boundary condition value, while the threshold for inlier set acceptance was set to a moderately high value (1m) in order to address possible calibration errors. The parameters did not require tuning.

The search region scale-factors in stereo verification algorithm (\cref{sec:stereoDetection}) were experimentally determined on a separate training set and were set to $\theta_{\mathrm{S1}} = 1.2$ and $\theta_{\mathrm{S2}} = 3.0$, while the NCC acceptance threshold was set to a high, conservative value $\theta_{\mathrm{ncc}} = 0.95$, and was not fine tuned.

\subsection{Comparison to state of the art: monocular setup}
\label{sec:mono_eval}
This section analyzes the performance of our IMU-assisted semantic segmentation method (ISSM), from \cref{sec:algDesc}, on the task of monocular obstacle detection. Our approach is compared to the SSM~\cite{KristanCYB2015}, which is currently the most recent state of the art for segmentation-based obstacle detection in USV.

We use the same performance measures as \cite{KristanCYB2015}. The accuracy of the sea-edge estimation is measured by mean-squared error and standard deviation over all sequences ($\mu_\mathrm{edg},\sigma_\mathrm{edg}$). The accuracy of obstacle detection is measured by number of true positives (TP), false positives (FP), and false negatives (FN), by F-measure, and the number of false positives per frame ($\alpha\mathrm{FP}$).

\begin{figure}
	\centering 
	\includegraphics[width=1\columnwidth]{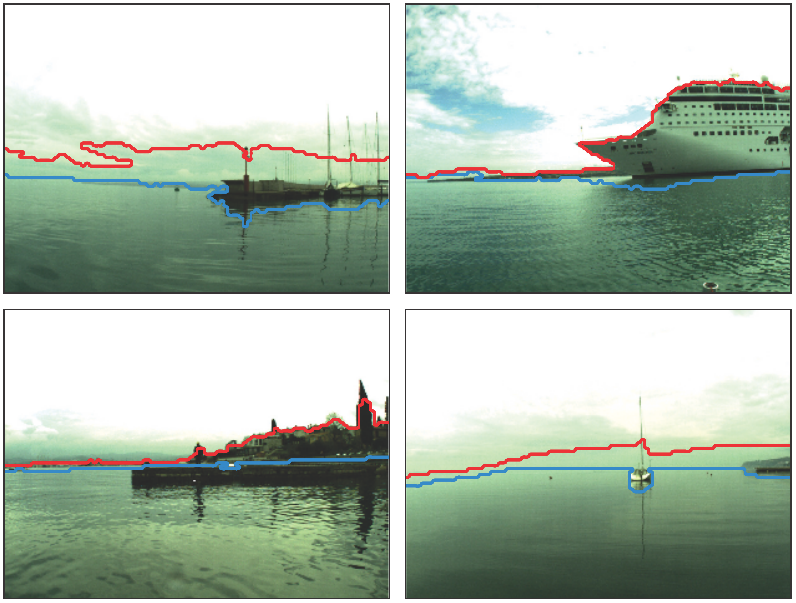}
	\caption{Qualitative comparison of ISSM (blue line) and original SSM (red line) on the task of sea-edge estimation using monocular camera. }
	\label{fig:comparisonSSM}
\end{figure}

The results of comparison are summarized in the first two rows of \cref{tab:allResults}. On the task of water-edge estimation, the proposed ISSM outperforms the original SSM
by \SI{29.8}{\%}. The improvement is statistically significant according to the Student's T-test~\cite{Student1908} with a confidence level of \SI{95}{\%}. Our ISSM also reduces
the false negative rate by \SI{66.9}{\%}, while drastically improving the number of true positive detections,
by \SI{61.3}{\%}. This is reflected in the F-score, which is improved by \SI{44.9}{\%}.

\begin{figure}
	\centering 
	\includegraphics[width=1\columnwidth]{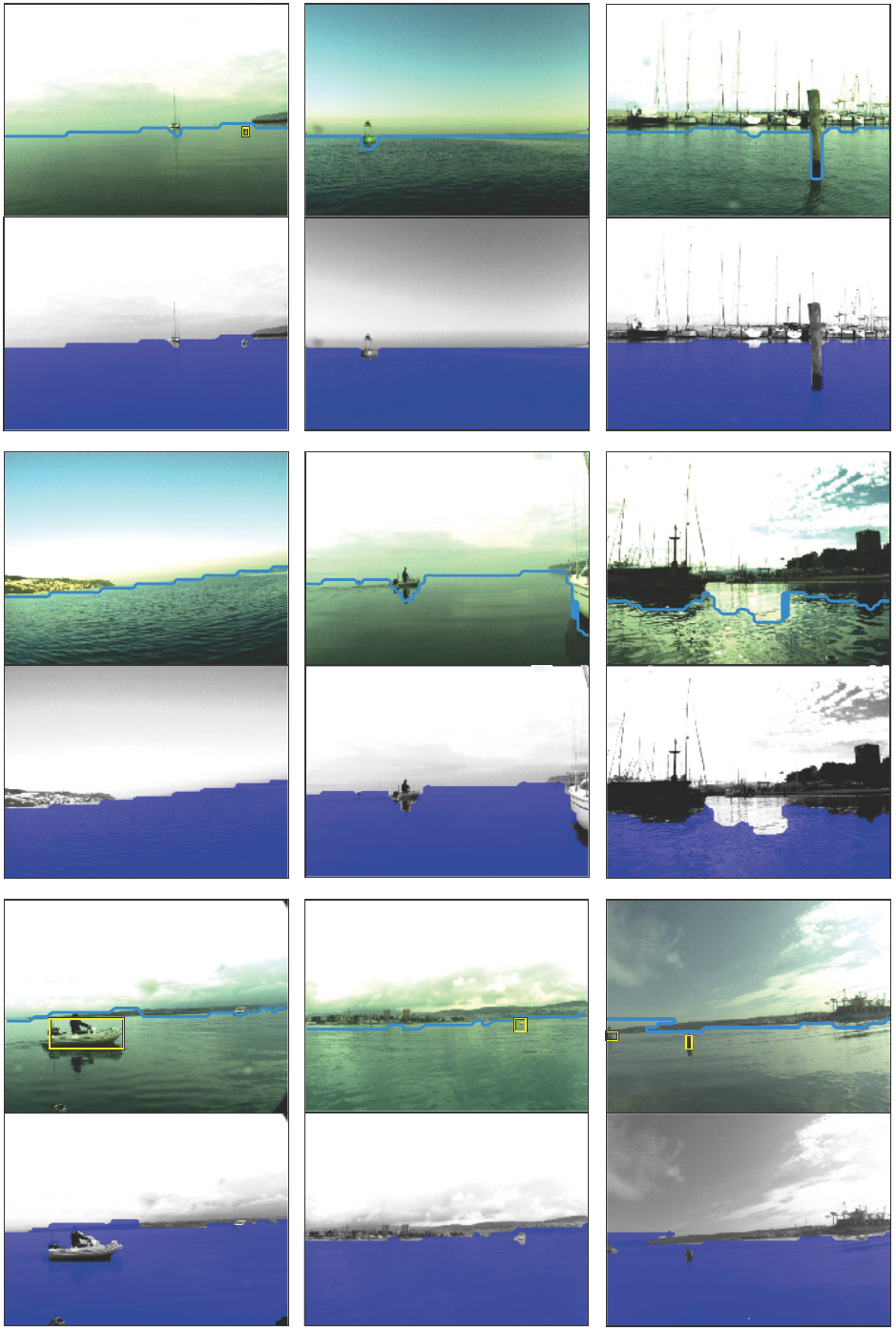}
	\caption{Qualitative examples of water segmentation for monocular camera setup. For each input image we separately show detected edge of the sea (blue), detected obstacles (yellow rectangle) and the portion of image segmented as water (blue).}
	\label{fig:qualitativeWaterObstacleDetection}
\end{figure}

\Cref{fig:comparisonSSM} shows a qualitative comparison of sea-edge detection of both algorithms. We can observe that the original SSM often over-estimates the extent of water. This leads to dangerous situations where obstacles are mistaken for water.
Additional examples of water segmentation by ISSM are illustrated in \cref{fig:qualitativeWaterObstacleDetection}. Our ISSM is able to correctly segment the images even in harsh conditions where the horizon is blurred.

\begin{figure}
	\centering 
    \includegraphics[width=1\columnwidth]{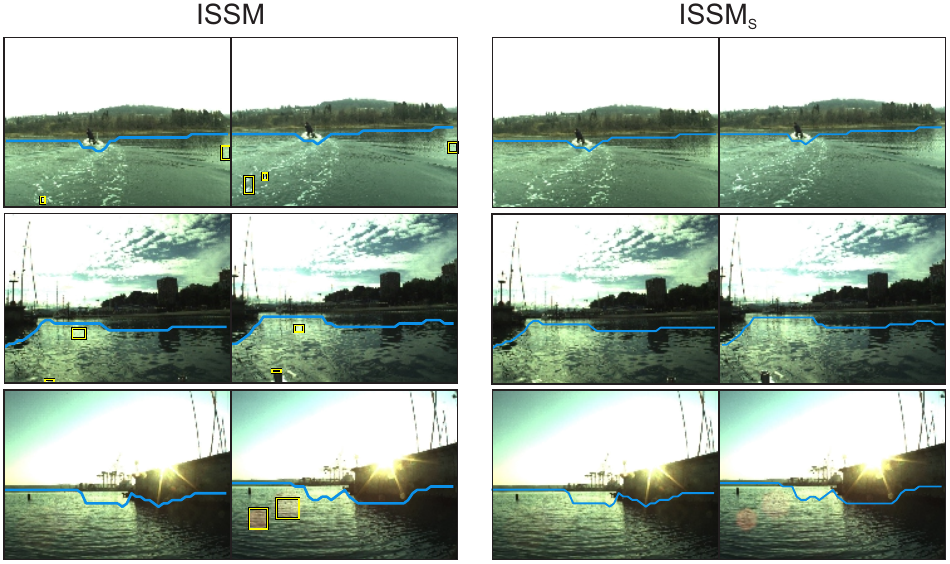}
    \caption{Qualitative comparison of ISSM with and without stereo verification (\cref{sec:stereoDetection}) in obstacle detection. The detected sea edge is denoted with blue line and detected obstacles with yellow rectangles. }
	\label{fig:comparisonSSM_obstacles}
\end{figure}

\subsection{Ablation study of the ISSM}\label{sec:ablationstudy}
An ablation study was performed to determine how each of our contributions from \cref{sec:algDesc} affects the detection performance. The following variants of ISSM were tested: (i) ISSM\textsubscript{M1}, that does not use modified hyper-priors (\cref{sec:hyperPriors}), (ii) ISSM\textsubscript{M2}, that does not use modified hyper-priors, nor water-component conditional-prior distribution (i.e., ${p(x_i = k \mid \boldsymbol{h}_t)}$ from \cref{sec:maskIMU} set to one for ${k = 3}$), (iii) ISSM\textsubscript{M3}, that does not use modified hyper-priors, nor sky-component conditional-prior distribution (i.e., ${p(x_i = k \mid \boldsymbol{h}_t)}$ set to one for ${k = 1}$), and (iv) ISSM\textsubscript{H}, that does not use conditional-prior distributions (i.e., ${p(x_i = k \mid \boldsymbol{h}_t)}$ from \cref{sec:maskIMU} set to one for all $k$). The results of variants are given in \cref{tab:allResults} (rows three to six).

The results of ISSM\textsubscript{M1} variant indicate a drastic improvement in sea-edge estimation (by \SI{29.3}{\percent}) compared to the original SSM as well as an improvement of \SI{42.5}{\percent} in F-score.
The ISSM\textsubscript{M2} variant yields an improvement in F-score (by \SI{26.4}{\percent}) compared to the original SSM, but its sea-edge estimation accuracy is reduced by \SI{5.6}{\%}. Among other variants in the group, this one has the lowest performance regarding sea-edge estimation and true-positive detections. The ISSM\textsubscript{M3} improves the sea-edge estimation by \SI{31.0}{\percent} and F-score by \SI{44.0}{\percent} compared to the original SSM. The ISSM\textsubscript{M3} outperforms other variants in the group on the task of sea-edge estimation as well as F-score measure, but it detects less true positives than the ISSM\textsubscript{M1}. The ISSM\textsubscript{H} variant improves the sea-edge estimation by \SI{6.0}{\%} and the F-score measure by \SI{23.9}{\%}. Among other variants in the group, this one achieves the lowest F-score results.

We observe that each of the described variants outperforms the original SSM in sea-edge estimation error as well as F-score. 
The variant ISSM\textsubscript{M3} achieves the best sea-edge estimation results as well as the best F-score. The best overall performance is obtained by combination of all variants (the proposed ISSM).

\begin{table}
\centering
\caption{
Comparison of our ISSM and ISSM\textsubscript{S} to SSM~\cite{KristanCYB2015} and SSM\textsubscript{S}.
We report sea-edge estimation error ($\mu_{\mathrm{edg}}$) and its standard deviation in brackets divided by height of the image to remove the effect of resolution, numbers of 
true positive (TP), false positive (FP), and false negative (FN) detections, the F-measure (F-score), and the average number of false positives per 
frame ($\alpha$FP).} \label{tab:allResults}
\input{table_all_results}
\end{table}

\subsection{Comparison to state of the art: stereo verification}
\label{sec:stereo_eval}
This section analyzes the performance of our IMU-assisted semantic segmentation algorithm with added stereo verification (ISSM\textsubscript{S}) from~\cref{sec:stereoDetection}. The original SSM~\cite{KristanCYB2015} is implemented strictly for monocular camera. For fair comparison, we have therefore also implemented a variant of SSM~\cite{KristanCYB2015} with added proposed stereo verification (\cref{sec:stereoDetection}), which we denote as SSM\textsubscript{S}. The results are summarized in \cref{tab:allResults} (rows seven to eight).

Comparing the results of SSM~\cite{KristanCYB2015} to its stereo extension SSM\textsubscript{S}, we observe a significant decrease of false positive detections (by \SI{90.9}{\percent}), which consequently boosts the F-score by \SI{35.7}{\percent}. Similarly, we observe a significant drop in false positive detections (by \SI{95.2}{\percent}) when comparing the results of ISSM and ISSM\textsubscript{S}. This leads to an improved F-score by \SI{46.5}{\percent}. From presented results, we conclude the proposed stereo verification step (\cref{sec:stereoDetection}) on average improves F-score by approximately \SI{41}{\percent}.

Comparing the results of SSM\textsubscript{S} and ISSM\textsubscript{S}, we observe that our ISSM\textsubscript{S} outperforms SSM\textsubscript{S} in every aspect. It detects \SI{65.2}{\percent} more true positives, has \SI{21.9}{\percent} fewer false positives, and \SI{59.7}{\percent} fewer false negatives, which consequently leads to higher F-score, boosted by \SI{54.2}{\percent}.

\Cref{fig:comparisonSSM_obstacles} shows a qualitative comparison of monocular and stereo obstacle detection. We observe that stereo verification step successfully removes false positive detections caused by sea foam (\cref{fig:comparisonSSM_obstacles} first row), small sun glitter (\cref{fig:comparisonSSM_obstacles} second row), and sun flares on camera lens (\cref{fig:comparisonSSM_obstacles} third row).

\subsection{Computational performance analysis}
\label{sec:speed}
We further studied how each of the contributions from \cref{sec:algDesc} affects the processing speed of the obstacle detection algorithm. We measure processing times of each ISSM variant, presented in~\cref{sec:mono_eval} and~\cref{sec:ablationstudy}. The impact of the proposed SSM modifications on the processing speed is summarized in \cref{tab:timesAlg}.

The computation and application of the IMU information does not additionally reduce speed of the original SSM~\cite{KristanCYB2015}, maintaining on average approximately $30$ % $69.87$
fps throughout the sequences, while drastically improving the performance both for sea edge estimation and obstacle detection.

Timings for stereo segmentation are gathered in~\cref{tab:times_alg_stereo}. Compared to its monocular counterpart, the stereo segmentation version is on average slower by approximately \SI{50}{\percent}, mainly due to processing of two images instead of one. The additional speed reduction, caused by obstacle verification (\cref{sec:stereoDetection}), depends on the number of detected obstacles and increases as more obstacles need to be verified.

Although we experience a decrease in frame rate in stereo setup, the cameras on our target USV do not support framerates above $10$ frames per second. This makes our proposed algorithm real-time, wile significantly improving the quality of obstacle detection by \SI{46.5}{\%}.

\begin{table}
\centering
\caption{Segmentation speed for original SSM~\cite{KristanCYB2015} and different variants of our ISSM. The $\Delta$t denotes time required for
processing a single frame, $\omega$ denotes the processing frame-rate.} \label{tab:timesAlg}
\input{table_timings}
\end{table}

\begin{table}
\centering
\caption{Measurement of segmentation and obstacle verification speed for SSM\textsubscript{S} and ISSM\textsubscript{S}. The $\Delta$t denotes time required for
processing a single frame, $\omega$ denotes the processing frame-rate.} \label{tab:times_alg_stereo}
\input{table_timings_stereo}
\end{table}

\subsection{Performance under extreme conditions}
\label{sec:extreme_conditions}
In this section, we compare the results obtained on sequences with three types of extreme conditions indicated in \cref{tab:datasetDetails}: (i) abrupt USV motion change, (ii) environment reflections in the water, and (iii) sun glitter. Note that some sequences contain multiple extreme conditions. The results are summarized in \cref{tab:extreme_conditions}.

\begin{figure}
	\centering
	\includegraphics[width=1\columnwidth]{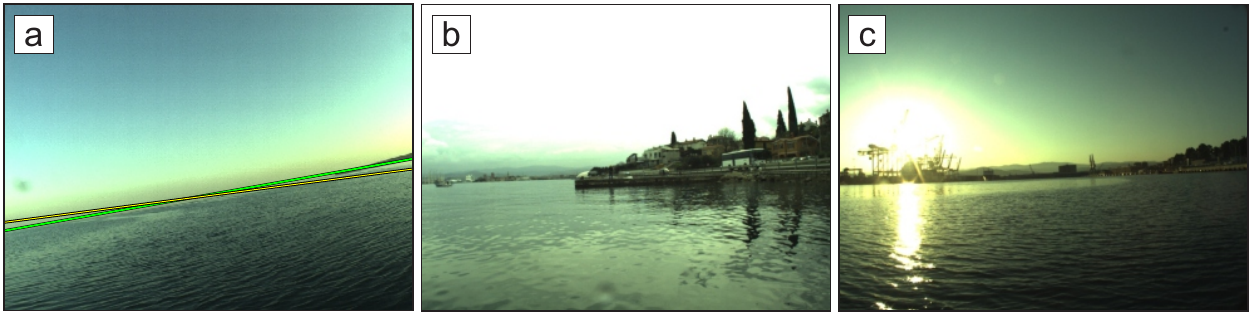}
	\caption{Different types of extreme cases. Image (a) shows abrupt motion change and estimated horizon lag where the ground truth horizon is denoted with green line and the estimated horizon is denoted with yellow line. Image (b) shows environment reflections in the water and image (c) shows sun glitter.}%
	\label{fig:extreme_cases}
\end{figure}

\subsubsection{Abrupt motion change} 
\label{sec:extreme_motion}
Sudden movement of the USV can be caused by strong waves and winds, or by acceleration/deceleration and abrupt steering of the USV. Due to internal filtering, the IMU readouts do not properly reflect the pose of the USV. As a result, the estimated horizon lags behind its true position (see \cref{fig:extreme_cases}a for example). This leads to reduced accuracy in conditional priors, which may affect the quality of the image segmentation. From the first four rows of \cref{tab:extreme_conditions}, we conclude ISSM outperforms the original SSM~\cite{KristanCYB2015} in sea edge estimation. Both algorithms detect the same amount of true positives and false negatives, however, the ISSM detects more false positives which leads to a lower F-score. By combining ISSM with stereo obstacle verification method from \cref{sec:stereoDetection} (ISSM\textsubscript{S}), the false positives are removed, while the number of true positives and false negatives are left unchanged, thus leading to F-score of $1$.

\subsubsection{Environment reflections} 
\label{sec:extreme_reflections}
The extent of environment reflections (boats, land, piers, buoys, etc.) increases with the flatness of the water surface, and affects the obstacle segmentation quality (see \cref{fig:extreme_cases}b). From the results in \cref{tab:extreme_conditions} (rows five to eight), we see that the ISSM outperforms SSM~\cite{KristanCYB2015} on the task of sea-edge estimation as well as F-score measure. The sea edge estimation is improved by \SI{46.8}{\percent}, which also leads to a better obstacle detection results (F-score improved by \SI{42.9}{\percent}). Stereo obstacle verification improves F-score by \SI{2.8}{\percent} when combining it with SSM, and by \SI{34.9}{\percent} when combining it with ISSM.

\begin{table}
	\centering
	\caption{Performance of algorithms under different types of extreme conditions.} \label{tab:extreme_conditions}
	\input{extreme_results}
\end{table}

\subsubsection{Glitter} 
\label{sec:extreme_glitter}
We distinguish between two types of sun glitter. The first type is caused by a low-lying sun, which creates a large region that significantly differs from the surrounding water. The second type are small local specular reflections caused by the sun. Both types of glitter can be seen in \cref{fig:extreme_cases}c. The small glitters cause small phantom detections in segmentation, while the large glitter causes mis-labeling of large water regions. Results from \cref{tab:extreme_conditions} (rows nine to twelve) show that  the proposed ISSM algorithm improves the sea edge estimation by as much as \SI{4.3}{\percent} compared to the SSM~\cite{KristanCYB2015}. The stereo obstacle verification method is capable of removing phantom detections caused by sun glitter and consequently improving the F-score by \SI{76.7}{\percent} when combining it with SSM and by \SI{76.5}{\percent} when combining it with ISSM.

\subsection{Failure cases} \label{sec:failure_cases}
\begin{figure*}
	\centering
	\includegraphics[width=1\textwidth]{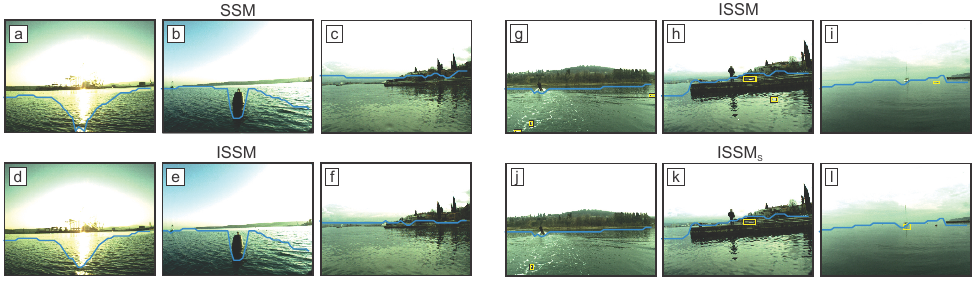}
	\caption{Qualitative examples of failure cases. The left side shows comparison between SSM~\cite{KristanCYB2015} and ISSM on the task of sea edge estimation under failure cases, while the right side shows comparison between ISSM and ISSM\textsubscript{S} on the task of obstacle detection under failure cases. The detected sea edge is denoted with blue line and detected obstacles are denoted with yellow rectangles.}%
	\label{fig:failure_cases}
\end{figure*}

Some sequences in the Modd 2 dataset pose a significant challenge for the segmentation. The failure cases are shown in \cref{fig:failure_cases}. The left side of \cref{fig:failure_cases} shows failures in sea edge estimation. \Cref{fig:failure_cases}(a,d) and \cref{fig:failure_cases}(b,e) show mis-labeling of water region due to sun glitter and opaque reflections. The water edge is estimated conservatively, preventing potentially dangerous false negatives near the boat. A potentially dangerous situation is shown in \cref{fig:failure_cases}(c,f), where a part of pier is labeled as water due to visual similarity. The ISSM provides a slightly better segmentation than SSM~\cite{KristanCYB2015}.

The right side of \cref{fig:failure_cases} shows failures in stereo obstacle verification. \Cref{fig:failure_cases}(g,j) shows false positives caused by sea foam. The ISSM detects false obstacles on the foam. The ISSM\textsubscript{S} discards some of the false positives, but one still remains because it is consistent in both views. \Cref{fig:failure_cases}(h,k) shows a dangerous mis-labeling of the pier as water, a true obstacle detection on part of the pier, and a false-positive detection due to sun glitter. The glitters are removed by ISSM\textsubscript{S}, but the detection on pier remains because it is visually consistent across the views. In the third example in \cref{fig:failure_cases}(i,l), the ISSM\textsubscript{S} verification incorrectly removed a true detection of the buoy because it appears very small at the observed distance.

%%%%%%%%%%%%%%%%%%%%%%
%%%   CONCLUSION   %%%
%%%%%%%%%%%%%%%%%%%%%%
\section{Conclusion}\label{sec:conclusionAndFutureWork} 
 We proposed a new segmentation model for obstacle detection in USVs. Our model extends the recent state-of-the-art semantic segmentation graphical model from~\cite{KristanCYB2015} by incorporating boat roll and pitch measurements from the on-board IMU. In addition, we propose a stereo verification scheme to reduce the false positive and false negative detections. To evaluate our new segmentation model, we captured a new challenging dataset that consists of several synchronized video sequences and IMU measurements, as well as annotations of water-edge and obstacles in each image. This new dataset is the largest multi-sensor USV dataset, and will be made publicly available.

A series of detailed experiments was conducted to analyze our proposed monocular algorithm (ISSM) and its stereo extension (ISSM\textsubscript{S}). Compared to the state-of-the-art SSM~\cite{KristanCYB2015}, the ISSM improves the sea edge estimation by \SI{29.8}{\percent} and the F-score by \SI{45.0}{\percent}. Our stereo verification approach boosts the F-score of the SSM~\cite{KristanCYB2015} and the ISSM by over \SI{35}{\%} and \SI{46}{\%}, respectively. The ISSM\textsubscript{S} outperforms the SSM~\cite{KristanCYB2015} with added stereo verification %(SSM\textsubscript{S}) 
by \SI{54.2}{\percent}. An ablation study showed that each part of our algorithm contributes to the performance, with best performance achieved with all parts combined.

The proposed ISSM and the stereo extension ISSM\textsubscript{S} significantly boost the performance in extreme conditions. Results show that our ISSM outperforms the SSM~\cite{KristanCYB2015} in water-edge estimation by \SI{35.5}{\percent} on average. In presence of glitter and mirroring, the ISSM outperforms the SSM~\cite{KristanCYB2015} by over \SI{27}{\percent} in F-score. Our stereo verification approach boosts the ISSM by over \SI{68}{\percent}. The stereo ISSM (ISSM\textsubscript{S}) outperforms the SSM~\cite{KristanCYB2015} with added stereo verification (SSM\textsubscript{S}) by approximately \SI{24}{\percent}.

Although the ISSM significantly outperforms the state-of-the-art SSM~\cite{KristanCYB2015}, it does so without significant drop in processing speed. The stereo extension reduces the speed by approximately \SI{50}{\percent} due to processing two images. However, the processing speed is still above the cameras frame rate, keeping the algorithm real-time.

Our future work will explore deeper integration of the stereo system into the graphical model, and extension with alternative camera modalities. Additional sensors, such as compass, clock, and GPS, could be used to determine the position of sun, and use it as a prior in explicit detection of reflections and glitter in the water (e.g.,~\cite{ahmed2011reflection}) to improve robustness. The current algorithm is able to adapt to the water appearance changes and runs in real-time, but uses very simple visual features. As part of our future work, we will explore feature learning to further improve the segmentation, while still keeping the algorithm real-time. The performance of our proposed method was evaluated offline on a desktop computer in Matlab environment. We plan to re-implement the ISSM to run on an on-board system of the USV and utilize the many related maritime datasets (e.g. PETS2016~\cite{patino2016pets}, MAR-DCT~\cite{bloisi2015argos}, SMD~\cite{Prasad2017}) for parameter tunning and additional evaluation.

%\addtolength{\textheight}{-12cm}
\section*{Acknowledgment}
\noindent
This work was in part supported by the Slovenian research agency ARRS programs P2-0214 and P2-0095 and the Slovenian research agency ARRS research project J2-8175. 
%
%%%%%%%%%%%%%%%%%%%%%%%%%%%%%%%%%%%%%%%%%%%%%%%%%%%%%%%%%%%%%%%%%%%%%%%%%%%%%%%%
% References
\section*{References}
\bibliographystyle{IEEEtran}
\bibliography{IEEEabrv,bibliography}
\end{document}

%% file: dataset_info.tex
\resizebox{\ifdim\width>\columnwidth\columnwidth\else\width\fi}{!}{%
\begin{tabular}{lrrccc}
\toprule
 & \multicolumn{1}{c}{\begin{tabular}[c]{@{}c@{}}$\mathrm{n_{frm}}$\end{tabular}} & \multicolumn{1}{c}{\begin{tabular}[c]{@{}c@{}}$\mathrm{n_{obj}}$\end{tabular}} & \begin{tabular}[c]{@{}c@{}}Sudden\\ movement\end{tabular} & \begin{tabular}[c]{@{}c@{}}Sun\\ glitter\end{tabular} & \begin{tabular}[c]{@{}c@{}}Land\\ reflection\end{tabular} \\
\midrule
Sequence 1  & 550 & 0.662 & \checkmark &  			& \checkmark \\
Sequence 2  & 470 & 0.731 &            &  			& \checkmark \\
Sequence 3  & 250 & 0     &            &  			& \checkmark \\
Sequence 4  & 240 & 0.629 &            &  			& \checkmark \\
Sequence 5  & 900 & 0.499 &            &  			& \checkmark \\
Sequence 6  & 820 & 2.520 &            &  			& \checkmark \\
Sequence 7  & 790 & 2.690 &            &  			& \checkmark \\
Sequence 8  & 670 & 0.890 &            &  			& \checkmark \\
Sequence 9  & 180 & 2.442 &            &  			& \checkmark \\
Sequence 10 & 110 & 0.996 &            &            &  			 \\
Sequence 11 & 280 & 0     & \checkmark & \checkmark &  			 \\
Sequence 12 & 200 & 0.230 &            & \checkmark &  			 \\
Sequence 13 & 210 & 0.945 &  		   & \checkmark &  			 \\
Sequence 14 & 175 & 1.337 &  		   & \checkmark & \checkmark \\
Sequence 15 & 415 & 1.231 &  		   &  			& \checkmark \\
Sequence 16 & 660 & 0.811 & 		   &	        & \checkmark \\
Sequence 17 & 310 & 0     &  		   &		    & 			 \\
Sequence 18 & 300 & 5.058 &  		   &		    & \checkmark \\
Sequence 19 & 520 & 1.972 &  		   &		    & 		     \\
Sequence 20 & 520 & 0.331 &  		   & \checkmark & 			 \\
Sequence 21 & 295 & 0.317 &  		   & \checkmark & 			 \\
Sequence 22 & 160 & 0.516 &  		   & \checkmark &  			 \\
Sequence 23 & 290 & 0.510 &  		   & \checkmark & \checkmark \\
Sequence 24 & 340 & 0.829 &  		   & \checkmark & 			 \\
Sequence 25 & 560 & 1.485 &  		   & \checkmark &		     \\
Sequence 26 & 170 & 4.018 &  		   & \checkmark & 			 \\
Sequence 27 & 620 & 0.732 &  		   & 		    & \checkmark \\
Sequence 28 & 670 & 0.855 & 		   & \checkmark & \checkmark \\
\bottomrule
\end{tabular}
}%

%% file: table_all_results.tex
\resizebox{\ifdim\width>\columnwidth\columnwidth\else\width\fi}{!}{%
\begin{tabular}{lcccccc}
\toprule
                       & $\mu_{\mathrm{edg}}$   & TP           & FP            & FN           & F-score        & $\alpha$FP     \\
\midrule
SSM                    & 0.084 (0.053)          & 264          & \textbf{1156} & 624          & 0.229          & \textbf{0.099} \\
ISSM                   & \textbf{0.059} (0.070) & \textbf{682} & 1708          & \textbf{206} & \textbf{0.416} & 0.146          \\
\midrule
%sea + ska mask
ISSM\textsubscript{M1} & 0.059 (0.071)          & \textbf{628} & 1643          & \textbf{260} & 0.398          & 0.141          \\
%skymask only
ISSM\textsubscript{M2} & 0.089 (0.066)          & 418          & \textbf{1385} & 470          & 0.311          & \textbf{0.119} \\ 
%sea mask only
ISSM\textsubscript{M3} & \textbf{0.058} (0.068) & 618          & 1513          & 270          & \textbf{0.409} & 0.130          \\
%modified hiper-priors only
ISSM\textsubscript{H}  & 0.079 (0.067)          & 441          & 1604          & 447          & 0.301          & 0.137          \\ 
\midrule
SSM\textsubscript{S}   & 0.084 (0.053)          & 215          & 105           & 673          & 0.356          & 0.009          \\
ISSM\textsubscript{S}  & \textbf{0.059} (0.070) & \textbf{617} & \textbf{82}   & \textbf{271} & \textbf{0.778} & \textbf{0.007} \\
\bottomrule
\end{tabular}
}%

%% file: table_timings.tex
\resizebox{\ifdim\width>\columnwidth\columnwidth\else\width\fi}{!}{%
\begin{tabular}{rcccccc}
\toprule
               & SSM   & ISSM\textsubscript{M1} & ISSM\textsubscript{M2} & ISSM\textsubscript{M3} & ISSM\textsubscript{H} & ISSM \\
\midrule
$\Delta$t [ms] & 29.34 & 29.75                  & 29.49                  & 29.44                  & 31.50                 & 34.38 \\
$\omega$ [fps] & 34.08 & 33.61                  & 33.91                  & 33.97                  & 31.75                 & 29.09 \\
\bottomrule
\end{tabular}
}%

%% file: table_timings_stereo.tex
\resizebox{\ifdim\width>\columnwidth\columnwidth\else\width\fi}{!}{%
\begin{tabular}{rcccccc}
\toprule
 & SSM\textsubscript{S} & SSM\textsubscript{S} & SSM\textsubscript{S} & ISSM\textsubscript{S} & ISSM\textsubscript{S} & ISSM\textsubscript{S}\\
 & seg.                 & ver.                 & seg. \& .ver         & seg.                  & ver.                  & seg. \& ver.         \\
\midrule
$\Delta$t [ms] & 58.69  & 21.99                & 80.68                & 68.77                 & 21.36                 & 90.13                \\
$\omega$ [fps] & 17.04  & 4.65                 & 12.39                & 14.54                 & 3.44                  & 11.10                \\
\bottomrule
\end{tabular}
}%

%% file: extreme_results.tex
\resizebox{\ifdim\width>\columnwidth\columnwidth\else\width\fi}{!}{%
\begin{tabular}{lcccccc}
\toprule
                      & $\mu_{\mathrm{edg}}$ [px] & TP  & FP   & FN  & F-score & $\alpha$FP \\
\midrule
\multicolumn{7}{c}{Abrupt motion change (\cref{sec:extreme_motion})}\\
\midrule
SSM     			  & 0.063 (0.043) 			  & 1   & 2    & 0   & 0.500   & 0.002 \\
ISSM   				  & 0.028 (0.022) 			  & 1   & 41   & 0   & 0.047   & 0.049 \\
SSM\textsubscript{S}  & 0.063 (0.043) 			  & 1   & 0    & 0   & 1.000   & 0.000 \\
ISSM\textsubscript{S} & 0.028 (0.022) 			  & 1   & 0    & 0   & 1.000   & 0.000 \\
\midrule
\multicolumn{7}{c}{Environment reflections (\cref{sec:extreme_reflections})}\\
\midrule
SSM     			  & 0.079 (0.044)             & 199 & 423  & 599 & 0.280   & 0.053 \\
ISSM   				  & 0.042 (0.042)             & 592 & 1027 & 206 & 0.490   & 0.128 \\
SSM\textsubscript{S}  & 0.079 (0.044) 			  & 150 & 92   & 648 & 0.288   & 0.012 \\
ISSM\textsubscript{S} & 0.042 (0.042) 			  & 527 & 74   & 271 & 0.753   & 0.009 \\
\midrule
\multicolumn{7}{c}{Glitter (\cref{sec:extreme_glitter})}\\
\midrule
SSM     			  & 0.116 (0.064) 			  & 110 & 886 & 32  & 0.193   & 0.229 \\
ISSM   				  & 0.111 (0.098)             & 137 & 975 & 5   & 0.219   & 0.252 \\
SSM\textsubscript{S}  & 0.116 (0.064)             & 110 & 13  & 32  & 0.830   & 0.003 \\
ISSM\textsubscript{S} & 0.111 (0.098)             & 136 & 14  & 6   & 0.932   & 0.004 \\
\bottomrule
\end{tabular}
}%